\pdfoutput=1

\documentclass[UTF8,11pt]{article}

\usepackage[]{EMNLP2022}

\usepackage{times}
\usepackage{latexsym}

\usepackage[T1]{fontenc}

\usepackage[utf8]{inputenc}

\usepackage{microtype}

\usepackage{inconsolata}

\usepackage{tabularx}
\usepackage{graphicx}
\usepackage{booktabs} 
\usepackage{fixltx2e} 
\usepackage{multirow}
\usepackage{makecell}
\usepackage{caption}
\usepackage{subcaption}
\usepackage{soul} 
\usepackage{amsmath}
\usepackage{listings}
\usepackage{colortbl}
\usepackage{xcolor}
\usepackage[normalem]{ulem}
\useunder{\uline}{\ul}{}
\usepackage{hyperref}

\usepackage{CJKutf8} 

\newcommand{\ours}{\texttt{GENIUS}}
\newcommand{\oursaug}{\texttt{GeniusAug}}
\title{\ours{}: Sketch-based Language Model Pre-training via Extreme and Selective Masking for Text Generation and Augmentation}

\author{Biyang Guo$^{1\dag}$, Yeyun Gong$^{2}$, Yelong Shen$^{3}$\\ {\bf Songqiao Han$^{1}$, Hailiang Huang$^{1*}$, Nan Duan$^{2*}$, Weizhu Chen$^{3*}$}
\\
        $^{1}$AI Lab, SIME, Shanghai University of Finance and Economics \\ 
        $^{2}$Microsoft Research Aisa, $^{3}$Microsoft Azure AI
\\
}

\begin{document}

\maketitle
\begingroup\def\thefootnote{$\dag$}\footnotetext{Work done during the internship at MSRA NLC group}\endgroup
\begingroup\def\thefootnote{$*$}\footnotetext{Corresponding author, email: hlhuang@shufe.edu.cn; nanduan@microsoft.com;wzchen@microsoft.com.}\endgroup
\begin{abstract}
We introduce \textbf{\ours{}}: a conditional text \textbf{\texttt{gen}}erat\textbf{\texttt{i}}on model \textbf{\texttt{u}}sing \textbf{\texttt{s}}ketches as input, which can fill in the missing contexts for a given sketch (key information consisting of textual spans, phrases, or words, concatenated by mask tokens). \ours{} is pre-trained on a large-scale textual corpus with a novel \textit{reconstruction from sketch} objective using an extreme and selective masking strategy, enabling it to generate diverse and high-quality texts given sketches. Comparison with other competitive conditional language models (CLMs) reveals the superiority of \ours{}'s text generation quality. 
We further show that \ours{} can be used as a strong and ready-to-use data augmentation tool for various natural language processing (NLP) tasks. Most existing textual data augmentation methods are either too conservative, by making small changes to the original text, or too aggressive, by creating entirely new samples. With \ours{}, we propose \textbf{\oursaug{}}, which first extracts the target-aware sketches from the original training set and then generates new samples based on the sketches.
Empirical experiments on 6 text classification datasets show that \oursaug{} significantly improves the models' performance in both in-distribution (ID) and out-of-distribution (OOD) settings. We also demonstrate the effectiveness of \oursaug{} on named entity recognition (NER) and machine reading comprehension (MRC) tasks.\footnote{Code and models are publicly available at \url{https://github.com/microsoft/SCGLab} and \url{https://github.com/beyondguo/genius}}


\end{abstract}


\section{Introduction}
When writing an article, we usually start by drawing up a general framework containing the key elements or thoughts we wish to convey (which we call as \textbf{sketch} in this work), on the basis of which we write the whole content. Motivated by this, we want to train a conditional language model to mimic this process to generate a full text based on a sketch, where the sketch may only make up a very small part of the full text. We define this as the \textit{sketch-based text generation}, which has rich potential applications such as human writing assistance \cite{Shih2019XLEditorPS}, automatic story generation \cite{yao2019plan-and-write}, or generating new samples as data augmentation for downstream NLP tasks \cite{cbert,PTM4aug}.

Sketch-based text generation can be viewed as the reconstruction of an \textbf{extremely-masked} text. Many pretrained transformer models (PTMs) such as BERT \cite{devlin2018bert}, RoBERTa \cite{liu2019roberta}, T5 \cite{2019t5-c4}, and BART \cite{lewis2020bart}, involve a text reconstruction objective during their pre-training. However, due to the low masking ratio (e.g. 15\% for BERT/RoBERTa/T5 and 30\% for BART), these models can only be used to recover a \textbf{mildly-masked} text, where a few tokens or small spans are missing. This makes them incompatible with our goal – to reconstruct the whole content based only on a sketch where most of the content may be corrupted.  \citet{zhu2019text-infilling,liu2019tigs,ilm} also propose text-infilling models, but these models are either based on traditional architectures like RNNs \cite{hochreiter1997lstm} or only trained on specific domains without large-scale pre-training, limiting their usage on downstream tasks.

In this work, we present a sketch-based text generation model – \textbf{\ours{}} (\textbf{\texttt{GEN}}erat\textbf{\texttt{I}}on \textbf{\texttt{U}}sing \textbf{\texttt{S}}ketch as input) – pretrained on a large-scale corpus using a novel \textit{reconstruction from sketch} pre-training objective with an \textbf{extreme} and \textbf{selective} masking strategy. Compared with previous related methods, \ours{} can generate more diverse, detailed, and coherent text based on sketches. \ours{} also exhibits a strong attribute controlling ability to generate content towards certain attributes like sentiment or topic, which makes \ours{} more flexible for conditional text generation and other applications.


We then illustrate the great potential of \ours{} for \textbf{textual data augmentation} as an important application of \ours{}. Data augmentation is widely used to enhance the generalization of deep learning models, especially when the training data is scarce or noisy \cite{da_survey}. Various data augmentation methods have been proposed in the NLP field, such as rule-based text-editing methods \cite{wei2019eda,feng-etal-2020-genaug, guo2022sta}, back-translation \cite{back_translation,back_translation1}, masked language model-based \cite{cbert,PTM4aug} and auto-regressive model-based methods \cite{lambada,PTM4aug}. However, we argue that most previous methods are \textit{either too conservative or too aggressive}: 1) Conservative methods like EDA \cite{wei2019eda}, MLM-based methods \cite{cbert,PTM4aug} only make small modifications to the original text. If the modifications are too large, they may harm the model's performance. Therefore, the generated samples are semantically and structurally very close to the original ones, which limits the diversity; 2) Aggressive methods aim to create completely new training samples rather than just altering the original ones. For example, LAMBADA \cite{lambada} fine-tunes GPT-2 \cite{GPT-2} to learn the patterns of training data and generate new sentences conditioned on certain labels. This introduces large diversity to the training set but is less controllable in terms of data quality, which means it has a higher chance of producing undesirable samples.

With this observation, we propose a sketch-based data augmentation method named \textbf{\oursaug{}} which balances between these two extremes, built upon the strong generation ability of \ours{}. With \oursaug{}, we first extract the essential parts of the text and then use the \ours{} model to produce new contexts around these essential parts. By doing so, the diversity is much higher than the conservative methods as larger parts of the original text are changed. At the same time, the quality is more reliable than aggressive methods since the core semantics are retained. Large-scale pre-training of \ours{} enables \oursaug{} to be a ready-to-use tool for augmentation without the need for further fine-tuning on downstream tasks (though we also show that fine-tuning may lead to extra performance gain). The flexible nature of \oursaug{} also makes it a general data augmentation method that can be easily applied to various NLP tasks, including text classification, named entity recognition, and machine reading comprehension. Extensive experiments show that \oursaug{} can generate high-quality training samples to boost the models' in-distribution (ID) and out-of-distribution (OOD) generalization performance, which outperforms traditional methods by large margins.

\begin{figure*}[t]
 \centering
\includegraphics[width=\textwidth]{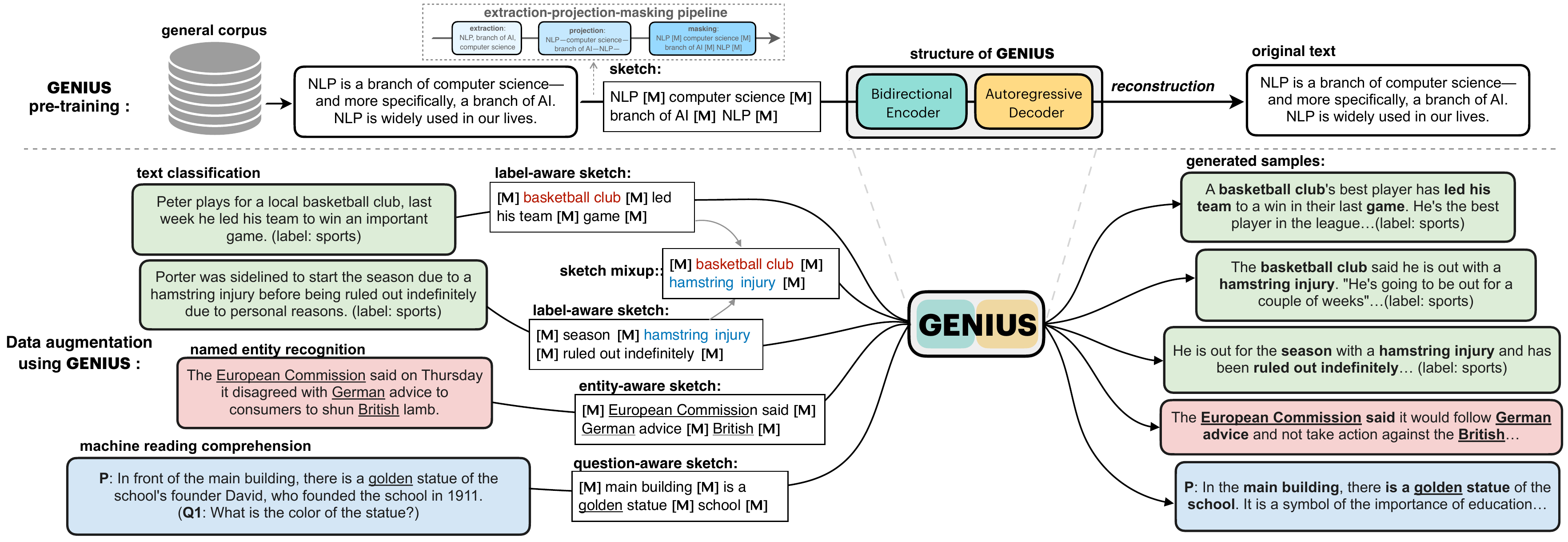} 
\caption{\small{Illustrations of the pre-training and augmentation process of \ours{}, where \texttt{[M]} refers to the model's mask token. The pre-training is conducted only once and then can be applied to various NLP tasks for data augmentation. During augmentation, we extract the \textit{target-aware} sketches as \ours{}'s inputs. The \textit{sketch mixup} shown in the figure is an example approach to further improve the diversity.}}
\label{sega-main-illustration}
\end{figure*}

The rest of the paper is organized as follows. Section \ref{sec:method} introduces the pre-training method of \ours{} and the method of applying \oursaug{} for data augmentation. Section \ref{sec:nlg-eval} discusses the performances of other possible pre-training strategies for \ours{} and evaluates the sketch-based text generation quality in comparison with other related methods. Section \ref{sec:exp-aug} elaborates on the data augmentation experiments including text classification, named entity recognition, and MRC tasks. We discuss the related work in Section \ref{sec:related-work} and the limitations of our work in Section \ref{sec:limitations}.

\section{Methodology}\label{sec:method}

\subsection{\ours{} Pre-Training}\label{sec:sega-pre-training}

We aim to train a conditional language model that can generate content from a sketch containing the key elements of the article. This is similar to the text reconstruction from corrupted text task, which is a common denoising pre-training objective for PTMs like BERT and BART (see Section \ref{sec:related-word-denoising}). However, these PTMs only reconstruct a small fraction of masked tokens from the original text (15\% for BERT and 30\% for BART), while we want to reconstruct the whole content from a sketch that may mask most of the text, which is beyond the capability of BERT or BART. Therefore, we propose a new pre-training task, \textit{reconstruction from sketch}, where we first extract a sketch from a document and then train the language model to recover the original text from the sketch.

To obtain a sketch that preserves the core semantics and the outline of the original text, we apply an \textit{extraction-projection-masking} procedure. We use the unsupervised keywords extraction algorithm YAKE \cite{campos2020yake} to extract keywords or key-phrases (up to 3-grams) from the original text, which account for roughly 1/5 of the text length. We then project these key parts back to the original text, allowing for multiple occurrences and overlaps of the key-phrases. We replace the rest of the text with a \textit{single mask token}, resulting in an average masking ratio of about 73\%\footnote{We sampled 1 million documents from our training set, the average masking ratio (\%) is $72.97 \pm 7.05$}. We construct about 27 million (27,599,676) \texttt{<sketch,text>} pairs from the C4-realnewslike corpus \cite{2019t5-c4} for pre-training. We use a bidirectional encoder and an auto-regressive decoder, initialized with BART weights. To facilitate future related research on other languages, we have also pre-trained a Chinese version of \ours{}, and is developing a multilingual version, see Appendix \ref{sec:pre-training} for more details. The upper part of Figure \ref{sega-main-illustration} shows an overview of the pre-training process.

The pre-training task of \ours{} differs from previous denoising pre-training objectives in two aspects: \textbf{Extreme masking}: We mask up to 80\% of the text, while previous methods usually mask a small proportion of text; \textbf{Selective masking}: We mask the less-informative parts of the text based on a sketch extraction pipeline, instead of masking random tokens/spans of the original text. 

 We will show that the following points are crucial for \ours{}'s pre-training: 1) the sketch should contain the key elements of the original text instead of randomly selected ones; 2) the order and occurrences of these elements should remain unchanged; 3) the missing parts are replaced with single mask tokens, which echos the text-infilling configuration of \ours{}'s backbone – BART. These designs alleviate the difficulty of high masking reconstruction, as well as the catastrophic forgetting problem \cite{french1999catastrophic} during the continued pre-training. Experiments and discussions about \ours{}'s sketch design are placed in Section \ref{sec:nlg-eval}.

\subsection{Data Augmentation with \ours{}}\label{sec:sega-augmentation}

Data augmentation is an important application for natural language generation (NLG) models, which is also a valuable evaluation of whether the generated text can be used in real applications. Here we introduce \textbf{\oursaug{}} which utilizes the strong ability of \ours{} for data augmentation.

\noindent\textbf{Target-aware sketch extraction.} To generate useful augmentation samples for downstream tasks using \ours{}, we need to feed the model with sketches that contain the information relevant to the task. We design a method called \textit{target-aware sketch extraction} to select the task-related parts from the original samples as the input sketches. Given a document $d$, all its n-grams (1 to 3-grams) $[w_{1},w_{2},...,w_{m}]$ and its \textit{target-related information} (\textbf{TRI}) $t$, we use a pre-trained encoder $\mathbf{E}$ to obtain their embeddings $e_{d}$, $[v_{1},v_{2},...,v_{m}]$, and $e_{t}$ respectively. The TRI depends on the task: for text classification, it is the category or its description; for NER, it is the entities; for machine reading comprehension, it is the question. We use sentence-BERT \cite{sentence-bert} as the encoder $\mathbf{E}$. Next, we compute the similarity between each n-gram and a fused embedding of the document and its TRI, which is defined as: \begin{align*}
    e_{f} &=  \lambda e_{d} + (1-\lambda)e_{t}\\
    s_{i} &= \frac{e_{f} \cdot v_{i}}{\| e_{f} \|  \| v_{i} \|}, i=1,2,...m
\end{align*}where $e_{f}$ is the fused embedding, $s_i$ is the similarity score for the $i$th n-gram $w_i$. We set $\lambda=0.5$ and choose the top 20\% n-grams with the highest similarity scores as the keywords/key-phrases. Finally, we construct the sketch from these selected parts following the steps in Section \ref{sec:sega-pre-training}.

\begin{table}[t]
\small
\setlength\tabcolsep{4.8pt}
\begin{tabularx}{\linewidth}{cX}
\toprule[1.5pt]
\textit{\textbf{sketch}}        & \scriptsize{\texttt{[M]} \textbf{use machine learning} \texttt{[M]} \textbf{AI techniques} \texttt{[M]}}           \\
\midrule[0.5pt]
\multirow{3}{*}{\textit{\ours{} output}}& \scriptsize{How do you \textbf{use machine learning} and other \textbf{AI techniques}? What are the benefits and disadvantages of AI?}                   \\
\midrule[0.5pt]
\multirow{4}{*}{\textit{\begin{tabular}[c]{@{}c@{}}prompt = \\ "\textbf{\textcolor{red}{Medicine}}"\end{tabular}}}     & \scriptsize{How do you \textbf{use machine learning} and \textbf{AI techniques} to help \textcolor{red}{\textbf{\textit{patients}}}? I am a software engineer. I have been working in AI for over 10 years. I am passionate about \textcolor{red}{\textbf{\textit{helping patients with their health problems.}}}}\\
\midrule[0.5pt]
\multirow{3}{*}{\textit{\begin{tabular}[c]{@{}c@{}}prompt = \\ "\textbf{\textcolor{red}{Finance}}"\end{tabular}}}      & \scriptsize{How do you \textbf{use machine learning} in your \textcolor{red}{\textbf{\textit{business}}}? \textbf{AI techniques} are a big part of the digital \textcolor{red}{\textbf{\textit{transformation of the economy.}}}}\\

\midrule[0.5pt]
\multirow{2}{*}{\textit{\begin{tabular}[c]{@{}c@{}}prompt = \\ "\textbf{\textcolor{red}{Good news}}"\end{tabular}}}      & \scriptsize{you \textcolor{red}{\textbf{\textit{can}}} now \textbf{use machine learning} and \textbf{AI techniques} to \textcolor{red}{\textbf{\textit{help you}}} get the most out of your new job.} \\
\midrule[0.5pt]
\multirow{3}{*}{\textit{\begin{tabular}[c]{@{}c@{}}prompt = \\ "\textbf{\textcolor{red}{Bad news}}"\end{tabular}}}      & \scriptsize{you \textcolor{red}{\textbf{\textit{can't}}} \textbf{use machine learning} to do what you want. \textcolor{red}{\textbf{\textit{It's not possible}}} to use \textbf{AI techniques} to predict what's going to happen in a particular situation.}\\
\bottomrule[1.5pt]
\end{tabularx}
\caption{\small{Examples of synthetic sentences generated by \ours{} and the effect of adding specific prompts for "attribute controlling". The \textcolor{red}{\textbf{\textit{red italicized}}} represents the text that is close to the given attribute.}}
  \label{tab:Text-examples}
\end{table}

\noindent\textbf{Generating new training samples.} After obtaining the sketch of a training sample, we use the pre-trained \ours{} to generate new samples conditioned on this sketch. We use beam search with random sampling to decode and generate the text. Table \ref{tab:Text-examples} illustrates an example sketch and its corresponding generated text by \ours{}. \ours{} is able to fill in the blanks (\texttt{[M]}) with multiple words or long spans, which is different from BERT or BART that can only fill in one or a few words. In the meanwhile, the key parts of the sketch remain in the generated text, which guarantees that the generated text won't have a large semantic shift from the original text. 

\noindent\textbf{Attribute controlling.} By adding a topic or sentiment prompt before the sketch, we can further control \ours{} to generate content towards certain attributes, as shown in the last four rows in Table \ref{tab:Text-examples}. Note that we didn't specifically pre-train or fine-tune \ours{} with an attribute-conditioned generation task, like CTRL \cite{keskar2019ctrl} or \cite{gpt-2-finetune}, nor did we use extra attribute models to control the generation like PPLM \cite{pplm}. The attribute controlling ability of \ours{} is acquired from the \textit{reconstruction from sketch} pre-training. This characteristic makes \ours{} flexible to control the quality and diversity during data augmentation, especially for sentiment or topic classification.

\noindent\textbf{More advanced options.} Apart from the standard usage introduced above, we find there exist other interesting ways to generate more diverse and high-quality training samples. For example, inspired by the Mixup technique \cite{zhang2018mixup}, we propose \textbf{sketch mixup} to \textit{combine the target-related parts from multiple training samples} to form the sketch, which is also shown in Figure \ref{sega-main-illustration}. Our experiments show this approach can bring further performance gains for some tasks compared with standard usages of \ours{}. This inspires us that the \textit{sketch designing} can be an interesting and worthy future research topic to further exploit the ability of the pre-trained \ours{} model.


\section{Experiments: Sketch-based Generation Pre-training Strategies and Evaluation}\label{sec:nlg-eval}


\subsection{Comparison of Different Strategies for \ours{}'s Pre-training}\label{sec:different-templates}
In Section \ref{sec:sega-pre-training} we introduce our default sketch design for \ours{}'s pre-training. Now we compare different possible sketch templates for pre-training: $T_1$ is a simple concatenation of the key elements joined by spaces, ordered by their importance given by the extractor. $T_2$ sorts the elements by their original order, and $T_3$ allows multiple occurrences and overlaps of key elements as in the original text. Our default sketch template described before is named $T_4$, which further replaces each of the missing parts with a single mask token. We also compare with $T_4$\textit{-random}, which extracts random n-grams for sketch construction as the only difference with $T_4$. Table \ref{tab:compare-sketch-types} shows examples of these templates.

 We pre-train \ours{} with these different sketch templates and evaluate their reconstruction performance on the dev set by ROUGE-1/2/L scores. The overall performance is ordered by $T_4$\textit{-random} $< T_1 < T_2 < T_3 < T_4$, which illustrates that our choice of $T_4$ helps \ours{} learn better during pre-training. Comparison between $T_2$ and $T_1$ shows the benefits of using the original order; comparison between $T_3$ and $T_2$ shows the benefits of keeping the original occurrences; comparison between $T_4$ and $T_3$ shows the importance of being consistent with BART by filling the missing parts with mask tokens; $T_4$\textit{-random} also possess these characteristics, but the random masking strategy makes model hard to learn in this high-masking setting, resulting in the lowest ROUGE scores among these choices. Therefore, we use $T_4$ as the default sketch template in this work.

\begin{table}[t]
\small
\centering
{\footnotesize
\begin{tabular}{lccc}
\toprule
\multicolumn{4}{l}{\textbf{\textit{passage}}:}\\
\multicolumn{4}{p{7cm}}{{\textbf{NLP} is a branch of \textbf{computer science}—and more specifically, a \textbf{branch of AI}. \textbf{NLP} is widely used in our lives.}} \\ \midrule
\multicolumn{4}{l}{\textit{\textbf{keywords/ key-phrases} (sorted by importance)}:}\\
\multicolumn{4}{l}{{[}NLP, branch of AI, computer science{]}}                              \\ \midrule
\multicolumn{4}{l}{$T_1$: NLP branch of AI computer science}                                                   \\ \midrule
\multicolumn{4}{l}{$T_2$: NLP computer science branch of AI}                                                    \\ \midrule
\multicolumn{4}{l}{$T_3$: NLP computer science branch of AI NLP}                                                 \\ \midrule
\multicolumn{4}{p{6.5cm}}{$T_4$ \textit{(default)}: NLP \texttt{\textbf{[M]}} computer science \texttt{\textbf{[M]}} branch of AI \texttt{\textbf{[M]}} NLP \texttt{\textbf{[M]}}}                       \\ \midrule
\multicolumn{4}{p{7cm}}{$T_{4}$\textit{-random  (extract random n-grams):}}         \\ 
\multicolumn{4}{p{7cm}}{\texttt{\textbf{[M]}} a branch \texttt{\textbf{[M]}} science \texttt{\textbf{[M]}} more specifically \texttt{\textbf{[M]}}}          \\
\bottomrule
\toprule
     Template               & \multicolumn{1}{c}{ROUGE-1}       & \multicolumn{1}{c}{ROUGE-2}       & \multicolumn{1}{c}{ROUGE-L}       \\ \midrule
$T_1$       & 28.32                             & 16.90                             & 24.05                             \\
$T_2$       & 28.56                             & 17.42                             & 26.41                             \\
$T_3$      & 28.70                              & 17.31                             & 26.52                             \\
$T_4$ \textit{(default)}       & \textbf{28.77}                    & \textbf{17.89}                    & \textbf{26.74}       \\ 
$T_4$\textit{-random}       & 17.82                             & 16.41                            & 17.78                             \\\bottomrule            
\end{tabular}}
\caption{\small{Comparison of different sketch templates for \ours{}'s pre-raining. \texttt{\textbf{[M]}} is the mask token.}}
\label{tab:compare-sketch-types}
\end{table}

\subsection{Quality Evaluation and Comparison for Sketch-based Text Generation}
Previous studies like blank infilling task \cite{shen2020blank} mainly evaluate how the reconstructed text \textit{resembles} the original one though CER \cite{morris2004CER} or BLEU \cite{Papineni02bleu:a} scores, the sketch-based text generation, however, is different since it favors more \textit{diversity} and there isn't a ground-truth to a sketch due to its high masking ratio. Therefore, we mainly evaluate how well the model can join the key elements in a fluent and informative way. 
We extract 1000 topic-related sketches from the HuffPost news dataset \cite{huffpost} from five topics (politics, sports, entertainment, tech, and business) using the same approach in Section \ref{sec:sega-augmentation}, and evaluate the sketch-based generation using the following metrics: \textbf{perplexity} \cite{jelinek1977perplexity} given by GPT-2 to measure the text fluency; \textbf{clf-error}, inspired by the Inception Score \cite{barratt2018note-IS} used for image generation evaluation, we trained a BERT-based classifier on 50K HuffPost news from the same topics as a scorer to measure how well the generated text can represent the original topic. The classification error is reported for this metric; \textbf{sketch-lost} which measures how the original sketch is retained in the generated text, calculated by the average of both word-level and fragment-level missing percentage; \textbf{recall} measures how much n-grams from the original text are restored in the generated text. We report the average recall of unigram, bigram, and the longest common subsequence levels; \textbf{diversity} is calculated by the percentage of new words introduced in the generated samples compared with the original samples. We also report the relative \textbf{length} of the generated text compared with the original text.
\begin{table*}[]
\centering
\small
\setlength\tabcolsep{4pt}
\begin{tabular}{lcccccc}
\toprule
              & \textbf{Perplexity(↓)} & \textbf{Clf-error(↓)} & \textbf{Sketch-lost(↓)} & \textbf{Recall(↑)} & \textbf{Diversity(↑)} & \textbf{Length} \\
              \midrule
\textit{raw sketch}    & 218.22                 & 9.32                  & 0.00                    & 20.21              & 0.00                  & 0.25            \\ \midrule
\textit{BART-large-infill} \cite{lewis2020bart}          & 56.04                  & 7.49                  & 0.96                    & 22.41              & 5.37                  & 0.37            \\
\textit{T5-CommonGen} \cite{gehrmann2021gem}         & 43.24                  & 12.73                 & 41.57                   & 15.06              & 8.48                  & 0.43            \\
\textit{CBART-large} \cite{cbart}         & 50.58                  & 10.08                 & 7.88                    & 22.32              & 8.07                  & 0.56            \\
\textit{ILM-ngram} \cite{ilm}     & 90.25                  & 12.2                  & \textbf{0.03}           & 24.71              & \textbf{30.58}                 & 1.47            \\ \midrule
\ours-T1-base   & 26.13                  & 8.97                  & 10.66                   & 24.71              & 8.59                  & 0.73            \\
\ours-T2-base   & 27.63                  & 9.08                  & 7.43                    & 22.77              & 6.55                  & 0.56            \\
\ours-T3-base   & 24.95                  & 8.20                  & 1.99                    & 24.71              & 9.16                  & 0.72            \\
\ours-T4-base            & 19.32                  & 8.12                  & 0.83                    & 25.35              & 9.23                  & 0.77            \\
\textbf{\ours{}} \scriptsize{(\ours-T4-large)} & \textbf{18.09}         & \textbf{7.09}         & 0.69                    & \textbf{29.51}     & 21.18                 & 1.71  \\
\bottomrule
\end{tabular}
\caption{\small{Evaluation for the sketch-based text generation using different methods. All models are not fine-tuned on the evaluation dataset. \ours{} generates more fluent and semantic-preserving text than other methods while also with high diversity.}}
\label{tab:nlg-eval}
\end{table*}

\begin{figure*}[t]
    \includegraphics[width=\textwidth]{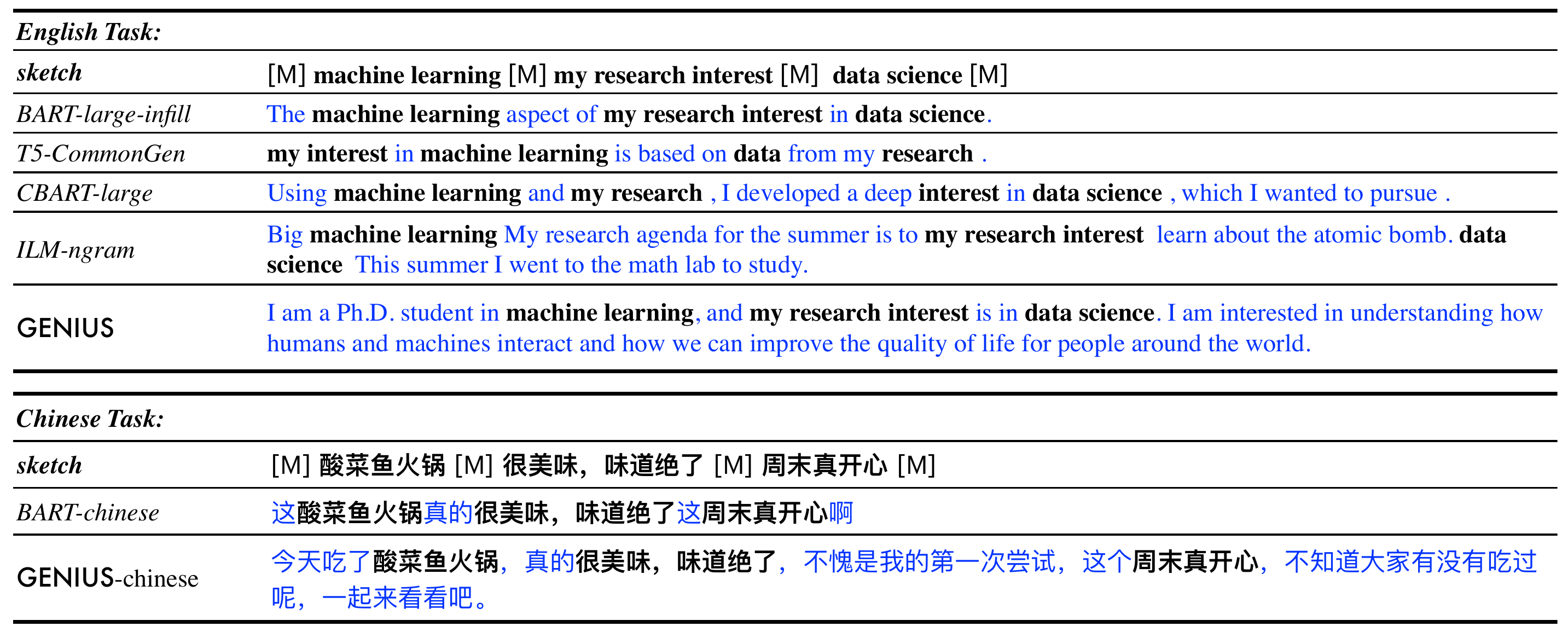}
    \caption{\small{Examples of different methods for sketch-based text generation. We illustrate both the English (default) and Chinese versions of \ours{}. The \textbf{bold} represents the key fragments from the sketch, while the \textcolor{blue}{blue} represents the newly generated contexts.}}
  \label{fig:nlg-examples}
\end{figure*}

We compare with the following existing models: \textbf{BART-large-infill} directly utilized the text-infilling ability of BART-large \cite{lewis2020bart} for generation; \textbf{T5-CommonGen} \cite{gehrmann2021gem} trains the T5 \cite{2019t5-c4} on the CommonGen \cite{lin2020commongen} dataset for keywords-to-text generation; \textbf{CBART-large} \cite{cbart} is a lexically constrained generation model that can gradually generate complete sentences given some keywords; \textbf{ILM-ngram} \cite{ilm} is the ngram infilling version of \textbf{ILM} which formulates the training sequences into a special blank-infilling structure and trains the GPT-2 model to fill the blanks. We also evaluate the performances of \ours{} using different sketch templates, denoted as \textbf{\ours{}-Tx-base}. Considering the superiority of $T_4$ discussed in Section \ref{sec:different-templates}, we pre-train a larger version using $T_4$ initialized with BART-large with longer training sequences, denoted as \textbf{\ours{}-T4-large}, which is also the default version of \ours{} in this work. Note that all models including \ours{} are not further fine-tuned on the HuffPost dataset. We use the publicly available model checkpoints of previous methods for evaluation.

According to the results in Table \ref{tab:nlg-eval}, \ours{} achieves the lowest perplexity, recall and clf-error compared with others. This means \ours{} can generate more fluent and semantic-preserving samples than other methods. These extracted sketches from the 5-class HuffPost dataset can be viewed as a corrupted training set where only some key information remains. \ours{} helps to reconstruct the corrupted training set and achieves the lowest classification error. The generated samples from \ours{} are also more diverse than other methods, except ILM-ngram. However, ILM-ngram has the highest perplexity and high clf-error. ILM-ngram is only pre-trained on 100K book texts, which partly explains the poor performance on a new domain. ILM uses a 15\% masking ratio during training, which also limits its ability in this high-masking reconstruction task. T5-CommonGen and CBART are limited to a few unigrams as input, and the output is limited to short sentences, resulting in high perplexity and clf-score when inputting sketches which may consist of multi-granularity elements. The comparison of different \ours{}-Tx-base models echos the results in Section \ref{sec:different-templates} that our default sketch template $T_4$ is superior to other templates, with better scores in all metrics. \ours{}-T4-large further improves the fluency and diversity by using larger model weights and longer training sequences.

Figure \ref{fig:nlg-examples} shows some generated examples by different models, for both English and Chinese tasks. One noticeable difference between \ours{} and previous methods is that \ours{} can generate longer text with more details. This feature is inherited from \ours{}'s extreme-masking pre-training where the model is asked to reconstruct large parts of the original text. ILM also generates longer text but is quite in-fluent. T5-CommonGen and CBART can generate relatively fluent text but may destroy the structure of the input sketch.

\begin{table*}[]
\small
\centering
\setlength\tabcolsep{4.8pt}
\begin{tabular}{l|cccccc>{\columncolor[gray]{0.95}}>{\itshape}c|cccc>{\columncolor[gray]{0.959}}>{\itshape}c}
                 & \multicolumn{7}{c|}{\textit{\textbf{ID evaluation}}}                                                                                           & \multicolumn{5}{c}{\textit{\textbf{OOD evaluation}}}                                                    \\ \midrule
         \textbf{Method} & \textbf{Huff}             & \textbf{BBC}              & \textbf{Yahoo}            & \textbf{20NG}             & \textbf{IMDB}             & \textbf{SST2}             & \textit{avg.}             & \textbf{H}$\Rightarrow$\textbf{B}            & \textbf{B}$\Rightarrow$\textbf{H}             & \textbf{I}$\Rightarrow$\textbf{S}             & \textbf{S}$\Rightarrow$\textbf{I}             & \textit{avg.}             \\ \midrule
\textit{none}             & 79.17          & \textbf{96.16} & 45.77          & 46.67          & 77.87          & 76.67          & 70.39          & 62.32          & 62.00          & 74.37          & 73.11          & 67.95          \\ \midrule
\textit{EDA} \citeyearpar{wei2019eda}     & 79.20          & 95.11          & 45.10          & 46.15          & 77.88          & 75.52          & 69.83          &  \underline{67.48}          & 58.92          & 75.83          & 69.42          & 67.91          \\
\textit{BackTrans} \citeyearpar{back_translation1}   & \underline{80.48}          & 95.28          & 46.10          & 46.61          & 78.35          & 76.96          & 70.63          & \underline{67.75}          & \underline{63.10}          & 75.91          & 72.19          & 69.74          \\
\textit{MLM} \citeyearpar{PTM4aug}     & \underline{80.04}          & 96.07          & 45.35          & 46.53          & 75.73          & 76.61          & 70.06          & \underline{66.80}          & \underline{65.39}          & 73.66          & 73.06          & 69.73          \\
\textit{C-MLM} \citeyearpar{PTM4aug} *   & \underline{80.60}          & 96.13          & 45.40          & 46.36          & 77.31          & 76.91          & 70.45          & \underline{64.94}          & \textbf{\underline{67.80}} & 74.98          & 71.78          & 69.87          \\
\textit{LAMBADA} \citeyearpar{lambada} * & \underline{81.46}          & 93.74          & \textbf{\underline{50.49}} & \underline{47.72}          & 78.22          & \underline{78.31}          & 71.66          & \underline{68.57}          & 52.79          & 75.24          & \underline{76.04}          & 68.16          \\
\textit{STA} \citeyearpar{guo2022sta}     & \underline{80.74}          & 95.64          & \underline{46.96}          & \underline{47.27}          & 77.88          & \underline{77.80}          & 71.05          & \underline{71.39}          & \underline{64.82}          & 74.72          & 73.62          & 71.13          \\ \midrule
\textbf{\textit{\oursaug{}}} (Ours)   & \underline{81.43}          & 95.74          & \underline{49.60}          & \underline{50.38}          & \textbf{\underline{80.16}} & \underline{78.82}          & 72.68          & \underline{74.87}          & \underline{66.85}          & 76.02          & 74.76          & 73.13          \\
\textbf{\textit{\oursaug{}-f}} (Ours) *  & \textbf{\underline{81.82}} & 95.99          & \underline{50.42}          & \textbf{\underline{50.81}} & 79.40          & \textbf{\underline{80.57}} & \textbf{73.17} & \textbf{\underline{76.18}} & \underline{66.89}          & \textbf{\underline{77.45}} & \textbf{\underline{80.36}} &  \textbf{75.22} \\ \bottomrule
\end{tabular}
\caption{\small{In-distribution (ID) and out-of-distribution (OOD) evaluations of different augmentation methods, where H, B, I, and S stand for BBC, Huff, IMDB, and SST2 respectively. The stared methods (*) need fine-tuning on the downstream tasks. \underline{Underline} means significant improvements over the \textit{none} baseline with paired student’s t-test, p < 0.05.}}
  \label{tab:results-clf}
\end{table*}

\begin{figure*}[t]
 \centering
  \begin{subfigure}[b]{0.49\textwidth}
         \centering
         \includegraphics[width=\textwidth]{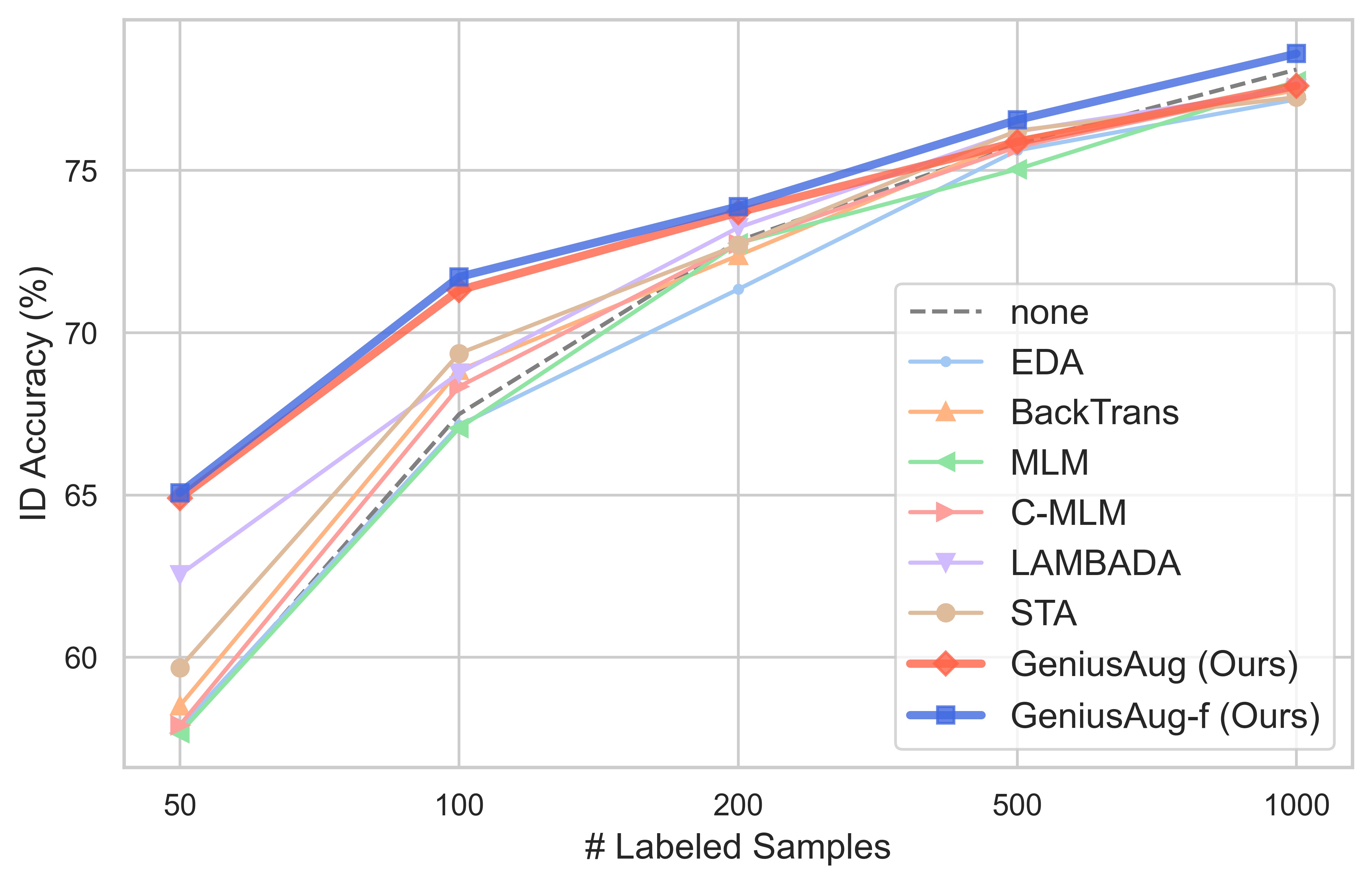}
         \caption{ID}
         \label{fig:ID-train-size}
     \end{subfigure}
\hfill
     \begin{subfigure}[b]{0.49\textwidth}
         \centering
         \includegraphics[width=\textwidth]{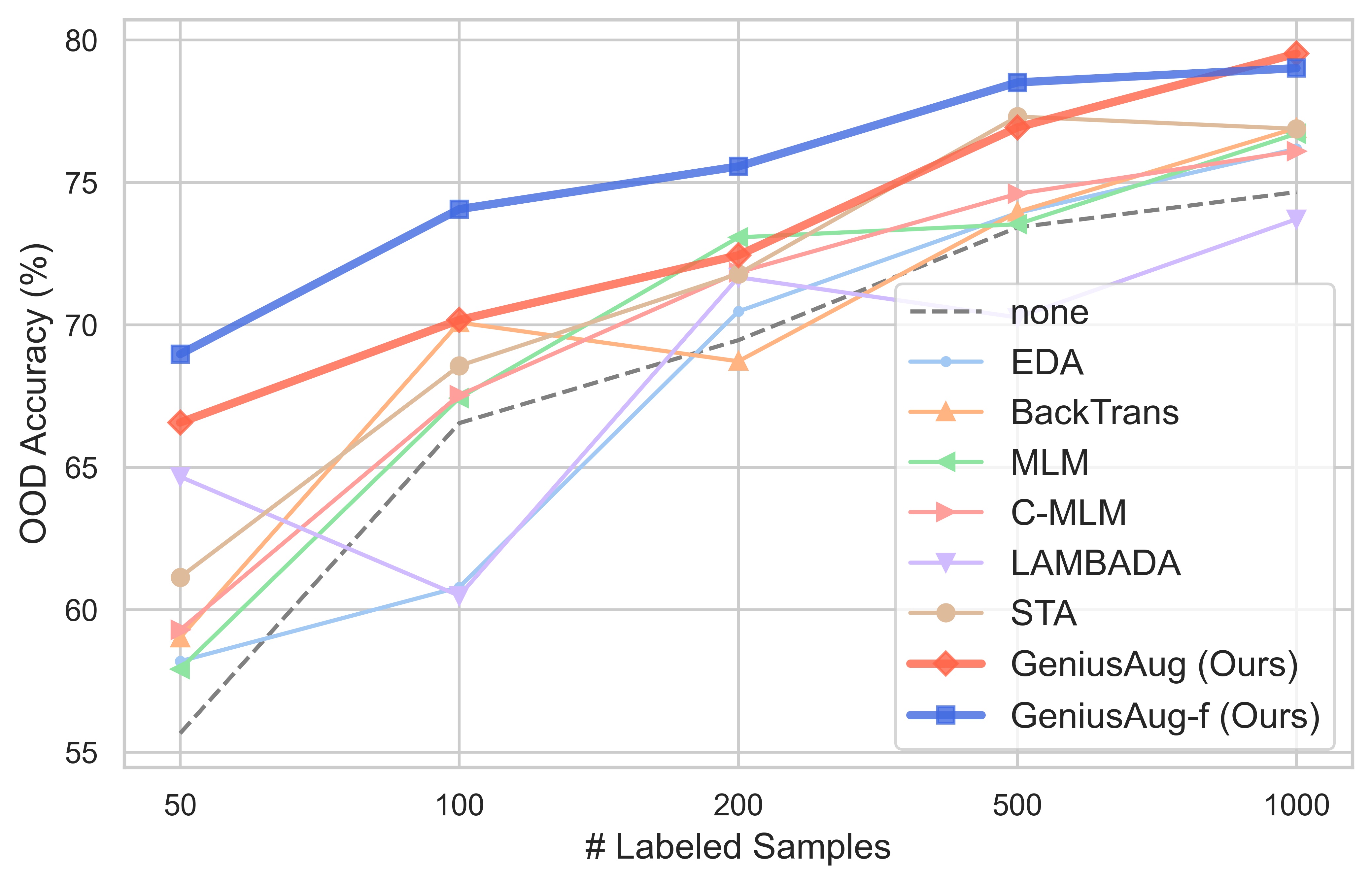}
         \caption{OOD}
         \label{fig:OOD-train-size}
     \end{subfigure}
 \caption{\small{Augmentation effectiveness when given different numbers of labeled samples in ID and OOD settings. We plot the averaged scores from all datasets.}}
 \label{train-size-id-ood}
\end{figure*}

\section{Experiments: Data Augmentation for Various NLP Tasks with \oursaug{}}\label{sec:exp-aug}
In this section, we will show that our proposed \oursaug{} method can effectively use sketch-based text generation for data augmentation, improving the downstream model performances for various NLP tasks, including text classification, NER, and MRC.

\subsection{Text Classification}\label{sec:exp-clf}
\subsubsection{Setup}
\textbf{Datasets.}\ \ 
We conduct experiments on 6 widely used datasets, including four topic classification datasets \textbf{BBC} \cite{data-bbc}, \textbf{Huff}\footnote{The original Huff dataset contains 41 categories. To facilitate the ID and OOD comparison, we only choose 5 categories that are the same as the BBC dataset.} \cite{huffpost}, \textbf{Yahoo} \cite{data-zhang-cnn}, \textbf{20NG}, and two sentiment classification datasets \textbf{SST2} \cite{Socher-sst2}, \textbf{IMDB} \cite{data-imdb}.
We experiment on a low-resource setting where $n = \{50,100,200,500,1000\}$ train/dev samples are randomly selected from the original train/dev sets of these datasets in our experiments. We use the original full test sets of these datasets for evaluation, which we call in-distribution (\textbf{ID}) evaluation. 
To further evaluate the model's generalization ability, we also design 4 groups of out-of-distribution (\textbf{OOD}) generalization tasks between the two movie review sentiment classification tasks – IMDB and SST2, and the two news classification tasks – BBC and Huff, following the experimental design in \cite{PTM-OOD}, where the model is first trained in one dataset and then directly evaluated on the other dataset without additional fine-tuning.

\noindent\textbf{Baselines.}\ \ 
The following augmentation methods are compared: rule-based \textbf{EDA} \cite{wei2019eda} and \textbf{STA} \cite{guo2022sta}; \textbf{BackTrans} \cite{back_translation2} uses the translation models from \cite{tiedemann-2020-tatoeba-translation-model} of four languages (de/ru/es/zh) in our experiments; \textbf{MLM} utilizes the masked language modeling (MLM) for words replacement; \textbf{C-MLM} \cite{PTM4aug} further fine-tunes a conditional MLM model by prepending the label to each sequence during MLM training. Note that \textbf{MLM} and \textbf{C-MLM} use BERT-base in their original work, while we use the stronger RoBERTa-large in our experiments; \textbf{LAMBADA} \cite{lambada} fin-tunes a conditional GPT-2 model to generate new samples by giving labels as prompts. 
Apart from directly using \textbf{\oursaug{}} for data augmentation, we also compare with a fine-tuned version \textbf{\oursaug{}-f} which is further fine-tuned on the downstream training set, where the model learns to reconstruct the original training sample given the target-aware sketch.
With each augmentation method, we scale up the training set to 2-5 times the original size and select the best model on the dev set for evaluation. All the augmentation methods are based on the same base text classifier using the same training hyper-parameters and model selection criteria (see Appendix \ref{sec:appendix-exp-model}). In the main experiments, the base classifier is \textit{DistilBERT-base} \cite{sanh2019distilbert}, which is an efficient lightweight Transformer model distilled from BERT. We also experiment on the stronger \textit{RoBERTa-large} \cite{liu2019roberta} classifier, discussed in the later part.

\subsubsection{Results}
 Table \ref{tab:results-clf} reports the averaged scores of $n=\{50,100,200,500,1000\}$ in both ID and OOD settings. The performances at each training size are shown in Figure \ref{train-size-id-ood}. 
 
\noindent\textbf{In-distribution evaluations.} Our proposed \oursaug{} and \oursaug{}-f both boosts the performance of base classifiers in the ID evaluations, with average improvements of around 2\% and 3\% respectively. \oursaug{} and \oursaug{}-f also outperform other baselines in most experiments by large margins. 
Among the baselines, rule-based STA and conditional generation-based LAMBADA are competitive methods, while other approaches bring only marginal gains or even degradation. Benefiting from a word roles recognition process and selective augmentation manner, STA can prevent the core semantics from being changed during augmentation while also introducing small perturbations to the original samples. However, rule-based operations may result in unnatural sentences and also limit the diversity. The strong reconstruction ability makes \oursaug{} superior to STA by generating more fluent and diverse samples.
LAMBADA can learn to generate diverse and coherent samples that belong to certain categories. However, since the generation is only conditioned on a label, the semantics of the generated text are more likely to be skewed. In comparison, \oursaug{} aims to generate more complementary contexts for a given sketch, thus better guaranteeing the quality of the generation.
  
\noindent\textbf{Out-of-distribution evaluations.}
OOD generalization is more challenging than ID evaluation since it requires the model to generalize to unknown distribution(s). In this setting, \oursaug{} and \oursaug{}-f bring much higher gains over non-augmentation baselines than in the ID evaluations, with average improvements of around 5\% and 7\% respectively. STA performs the best among the baselines thanks to its selective operations to protect core words and remove potential noise, which are helpful for generalization. LAMBADA exhibits severe degradation in the BBC$\Rightarrow$Huff generalization task. By checking the performances of LAMBADA at all training sizes, we find the OOD generalization performance of LAMBADA is even getting worse when more training data is provided. This phenomenon indicates that LAMBADA may have learned the dataset bias during fine-tuning on the source dataset, which harms the OOD generalization ability. The nature of \oursaug{} and \oursaug{}-f makes them very suitable for the OOD generalization: By masking the unimportant parts in the original text, \oursaug{} prevents the potential noise to be enhanced during generation; by keeping the key parts unchanged, the core semantics of the target is preserved during generation, reducing the risk of semantic drift.

\noindent\textbf{Performance at the different resource levels.}
Figure \ref{train-size-id-ood} shows the performances of different methods with different numbers of labeled training samples. In the severe low-resource scenarios ($n=\{50,100\}$), \oursaug{} and \oursaug{}-f are significantly stronger than other baselines. With labeled data getting richer, the gap between these methods is getting smaller in the ID setting, but \oursaug{} and \oursaug{}-f continue to maintain huge advantages over most baselines in the OOD setting.

\noindent\textbf{Effectiveness of fine-tuning.}
Compared with \oursaug{}, \oursaug{}-f achieves better results in most datasets. During \ours{}'s pre-training process, the model inputs are \textit{general} sketches, while in the fine-tuning procedure we are using \textit{target-aware} sketches, this helps the \ours{} to generate more target-related contents.

\begin{figure}[t]
 \centering
\includegraphics[width=\columnwidth]{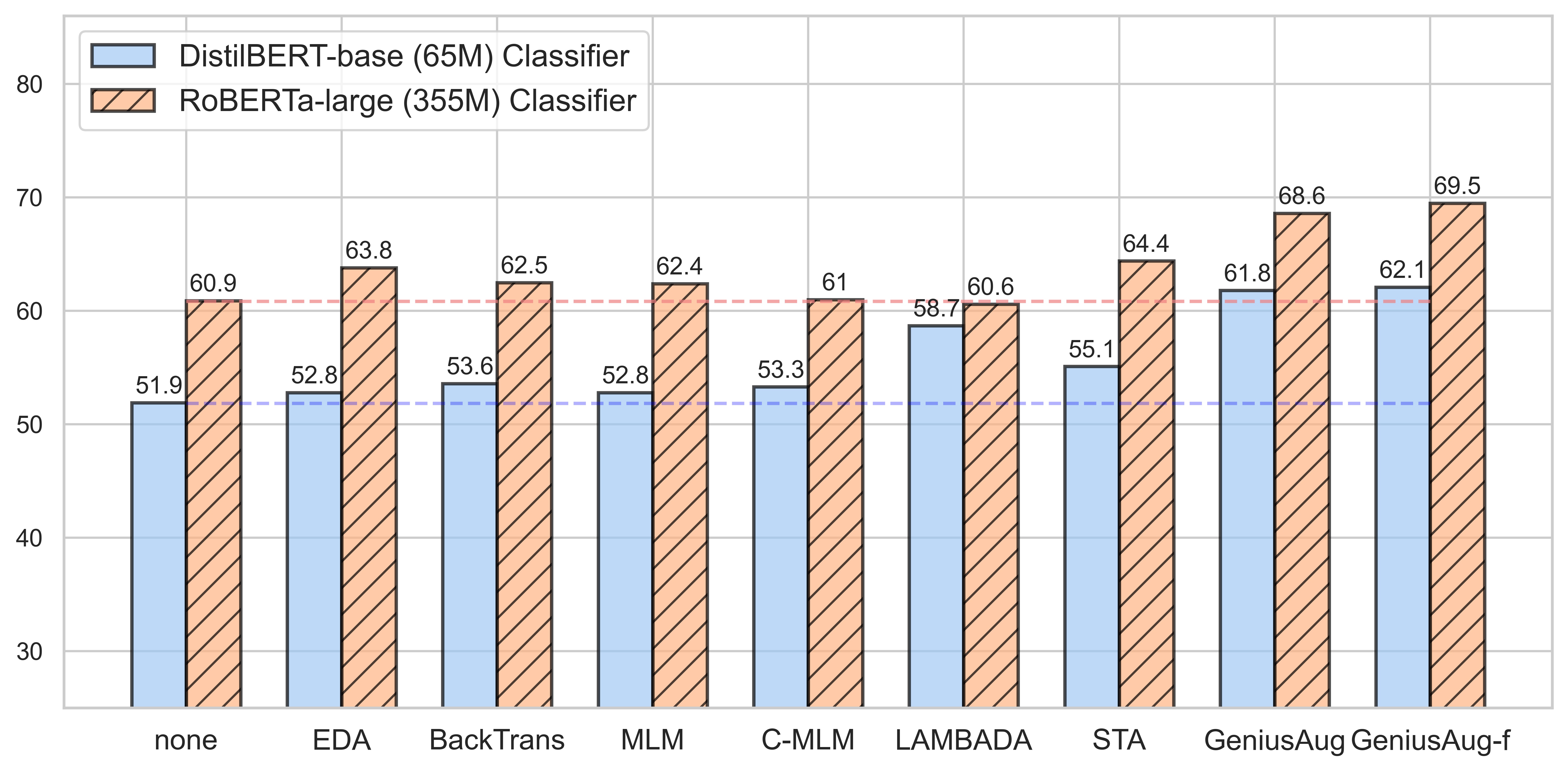}
\caption{\small{Performances on different classifiers – DistilBERT-base and a stronger RoBERTa-large. The results are averaged across all datasets on a $n=50$ setting.}}
\label{sega-distilbert-roberta}
\end{figure}

\noindent\textbf{Makes strong classifier stronger.}
In the above experiments, we use DistilBERT-base for the classifier, which has around 65 million parameters. Here we also evaluate the augmentation effectiveness using a much stronger RoBERTa-large classifier, with more than 355 million parameters. Due to the limited time/resources, we currently only experiment with $n=50$ for each dataset, as shown in Figure \ref{sega-distilbert-roberta}. Surprisingly, \oursaug{}/\oursaug{}-f \textit{helps the weaker DistilBERT-base classifiers to outperform the strong RoBERTa-large classifiers} (comparing the orange bar of "none" and blue bars of "\oursaug{}" and "\oursaug{}-f"). \oursaug{}/\oursaug{}-f also makes the RoBERTa-large classifiers stronger, improving their average accuracy scores by 8-9\% and outperforming other baselines by at least 4\%.

\noindent\textbf{Boosting the diversity via \textit{Sketch Mixup}.}
In the previous experiments, each sketch is extracted from a single training sample, which somehow is limited in diversity. Inspired by the Mixup \cite{zhang2018mixup}, we propose \textbf{\oursaug{}-mixup} to generate samples based on the mixed-up sketch that \textit{combines the target-related parts from multiple training samples}. By doing so, the generated samples exhibit larger differences from the original samples, while still being label-preserving. Experiments on all the datasets ($n=50$) show that \oursaug{}-mixup can further boost the performance for most experimented tasks. Note that \oursaug{}-mixup doesn't need a fine-tuning step. There are other possible ways to improve the diversity of \oursaug{}-generated dataset, such as applying synonyms replacement or changing the element order on the sketches, which we will explore in future work.

\begin{table}[t]
\small
\centering
\setlength\tabcolsep{3.5pt}
\begin{tabular}{c|cccc|c}
\toprule
   Dataset   & \textit{none}    & \textit{pre-sota} & \textit{GA}             & \textit{GA-f}           & \textit{\textbf{GA-mixup}}         \\
   \midrule
Huff  & 71.44 & 76.72  & 78.53          & 77.99 & \textbf{78.74} \\
BBC   & 94.94 & 95.60  & 94.90 & \textbf{95.30} & 95.14          \\
IMDB  & 58.78 & 62.57  & 68.74          & 65.54 & \textbf{69.58} \\
SST2  & 67.22 & 75.11  & 73.65 & \textbf{76.40} & 71.42          \\
Yahoo & 29.81 & 44.13  & 40.23          & 44.02 & \textbf{45.50} \\
20NG  & 23.78 & 26.06  & 33.42          & 31.15 & \textbf{38.77} \\
\midrule
avg.  & 57.66 & 63.37  & 64.91          & 65.07 & \textbf{66.53} \\ 
\bottomrule
\end{tabular}
\caption{\small{\texttt{GA} denotes \oursaug{}. \oursaug{}-mixup combines the key information from various samples to boost the diversity of synthetic data. The experiments are conducted at $n=50$ for each dataset.}}
  \label{tab:diversity-mixup}
\end{table}

\subsection{Augmentation for Other NLP Tasks}\label{sec:exp-ner-mrc}

\subsubsection{Setup}
We use the \textbf{CoNLL03} \cite{data-conll03} dataset for the named entity recognition (NER) task and \textbf{SQuAD} \cite{data-squad} for the machine reading comprehension (MRC) task. We sample the \textit{first} $n \in \{50,100,200,500\}$ labeled samples from the original datasets for our experiments. We use CoNLL03's original full test set for evaluation. For SQuAD, since the test set is not publicly available, we report the results on the development set. We report the F1 score for CoNLL03 and the exact match (EM) score for SQuAD. The base NER/MRC models are all based on \textit{BERT-base}.

For NER, Synonyms Replacement (\textbf{SR}), Mention Replacement (\textbf{MR}) \cite{simple-ner-aug}, \textbf{Rule-mix}\cite{simple-ner-aug}, and \textbf{MELM} \cite{zhou2022melm} are used as baselines. For MRC, \textbf{SR} and \textbf{BackTrans} are used. The other settings are the same as text classification tasks.
\begin{table}[t]
\small
\centering
\setlength\tabcolsep{3.5pt}
\begin{tabular}{l|cccc|c}
\toprule
                                  \multicolumn{6}{c}{\textbf{NER} (CoNLL03)}                                               \\ \midrule
$n$                  & \small{50}               & \small{100}              & \small{200}              & \small{500}       & \textit{avg.}       \\
\midrule
\textit{non-aug}     & 39.28          & 54.97          & 63.88          & 73.78     &   57.98   \\
\textit{SR} \citeyearpar{wei2019eda}      & 41.15 & 56.79 & 61.90 & 73.98   & 58.46       \\
\textit{MR} \citeyearpar{simple-ner-aug}       & 47.75 & 58.85 & 61.76 & 73.15     &   60.38   \\
\textit{Mix-rule} \citeyearpar{simple-ner-aug} & 45.78 & 55.56 & 61.60 & 72.55     &   58.87   \\
\textit{MELM} \citeyearpar{zhou2022melm}    & 46.45 & 53.05 & 60.96 & \textbf{77.36} & 59.46 \\
\midrule
\textit{\textbf{\oursaug{}} (Ours)} & \textbf{49.17} & \textbf{61.12} & \textbf{66.10} & 74.69 & \textbf{62.77}  \\
\bottomrule
\toprule
                       \multicolumn{6}{c}{\textbf{MRC} (SQuAD)}                                                                           \\ \midrule
$n$       & \multicolumn{1}{c}{\small{50}}          & \multicolumn{1}{c}{\small{100}}         & \multicolumn{1}{c}{\small{200}} & \multicolumn{1}{c}{\small{500}}         & \multicolumn{1}{|c}{\textit{avg.}}\\

\midrule
\textit{non-aug}   & 15.74 & 21.67    & 31.19     & 47.98     &  29.15    \\
\textit{SR} \citeyearpar{wei2019eda}    & 18.45   & 25.35     & 35.98  & 50.86        &    32.66     \\
\textit{BackTrans} \citeyearpar{back_translation1} & \textbf{19.26}  & 26.13 & 36.21   & 50.31         &    32.98    \\
\midrule
\textit{\textbf{\oursaug{}} (Ours)}    & 19.03   & \textbf{28.60}       & \textbf{37.02}       & \textbf{51.83}   &  \textbf{34.12}   \\
\bottomrule
\end{tabular}
\caption{\small{Data augmentation for the NER and MRC tasks with different labeled sizes.}}
  \label{tab:results-ner-mrc}
\end{table}

\subsubsection{Results}
\textbf{Named entity recognition.}
The upper part of Table \ref{tab:results-ner-mrc} summarizes the comparison for the CoNLL03 task with different sizes of labeled data. 
When the number of labeled data $n\in \{50,100,200\}$, \oursaug{} outperforms all the baselines by a large margin. When $n$ reaches $500$, \oursaug{} outperforms previous methods, except MELM. The rule-based methods SR, MR, and Mix-rule can improve the recognition performance when the labeled data size is small ($n\in\{50,100\}$). However, these methods may generate non-fluent sequences, or make the lexical pattern around entities unnatural, which  explains why these methods harm the performance when their training size becomes larger ($n\in\{200,500\}$). MELM utilizes a novel masked entity language modeling task to train the model for predicting new entities, which achieves the best F1 score when $n=500$. However, when the training size is small, MELM may not be able to train a satisfactory language model for generating proper new entities, which degrades the performance ($n\in\{100,200\}$). Compared with rule-based methods, \oursaug{} can generate more diverse and natural text, which is helpful for the NER model to learn new patterns for entity recognition. Compared with MELM, \oursaug{} doesn't rely on a fine-tuning process and thus can achieve better results when the training size is extremely small. A major difference between \oursaug{} and MELM is that \oursaug{} aims to diversify the context while MELM focuses on introducing more entities. We will explore the combination of \oursaug{} and MELM in future work to further improve the performance.

\noindent\textbf{Machine reading comprehension.}
 The lower part of Table \ref{tab:results-ner-mrc} shows the exact match (EM) scores for SQuAD in different resource levels.  \oursaug{} achieves the best results when $n=\{100,200,500\}$ and comparable scores with BackTrans when $n=50$. The two baselines SR and BackTrans are both effective augmentation methods among all training sizes. Compared with the baselines, \oursaug{} brings more diversity to the context around the answer, which helps improve the understanding ability of MRC models. There are also question data augmentation (QDA) methods specifically designed and pre-trained for MRC and QA tasks \cite{alberti2019synthetic, CRQDA}. We don't compare these methods in our experiments, since they need relatively large data for pre-training, which are not applicable in our low-resource setting. In addition, \oursaug{} is orthogonal to these QDA methods: \oursaug{} focuses on augmenting the context around the current answer, while QDA methods aim to generate new questions for the current context. Therefore, QDA methods can be combined with \oursaug{} to generate more diverse training samples for MRC and QA tasks.

\begin{table}[t]
\small
\centering
\setlength\tabcolsep{3.3pt}
\begin{tabular}{l|ccccc|l}
\toprule
\multicolumn{7}{c}{\textbf{ID (Huff)}}                                                                                                                                            \\ \midrule
n                                             & 50                   & 100                  & 200                  & 500                  & 1000                 & avg.     \\ \midrule
none                                          & 71.44              & 78.99              & 78.57              & 81.48              & 85.35              & 79.17 \\ \midrule
\oursaug{}                                          & \textbf{78.53}     & 79.97              & 80.94              & 82.65              & 85.05              & 81.43 \\
\multicolumn{1}{r|}{$-$ \textit{PT}} & 72.92              & 76.73              & 79.89              & 81.82              & 83.94              & 79.06 \\ \midrule
\oursaug{}-f                                        & 77.99              & \textbf{81.01}     & 81.00              & \textbf{83.14}     & 85.94              & \textbf{81.82} \\
\multicolumn{1}{r|}{$-$ \textit{PT}} & 69.85              & 77.32              & 79.87              & 81.61              & \textbf{85.98}     & 78.93 \\
\bottomrule
\toprule
\multicolumn{7}{c}{\textbf{OOD (Huff$\Rightarrow$BBC)}}                                                                                                                                           \\ \midrule
n                                             & 50                   & 100                  & 200                  & 500                  & 1000                 & avg.    \\ \midrule
none                                          & 39.62              & 63.58              & 65.20              & 65.90              & 77.30              & 62.32 \\ \midrule
\oursaug{}                                          & 66.14              & 74.94              & 70.84              & 78.46              & \textbf{83.98}     & 74.87 \\
\multicolumn{1}{r|}{$-$ \textit{PT}} & 40.70              & 64.28              & 62.38              & 72.22              & 78.58              & 63.63 \\ \midrule
\oursaug{}-f                                        & \textbf{66.70}     & \textbf{76.18}     & \textbf{77.24}     & \textbf{81.24}     & 79.56              & \textbf{76.18} \\
\multicolumn{1}{r|}{$-$ \textit{PT}} & 42.40              & 63.06              & 62.00              & 68.18              & 78.80              & 62.89 \\
\bottomrule
\end{tabular}
\caption{\small{Ablation study on the effectiveness of \ours{}'s large-scale pre-training for data augmentation}}
\label{ablation-pretrain}
\end{table}

\begin{figure}[h]
 \centering
\includegraphics[width=\columnwidth]{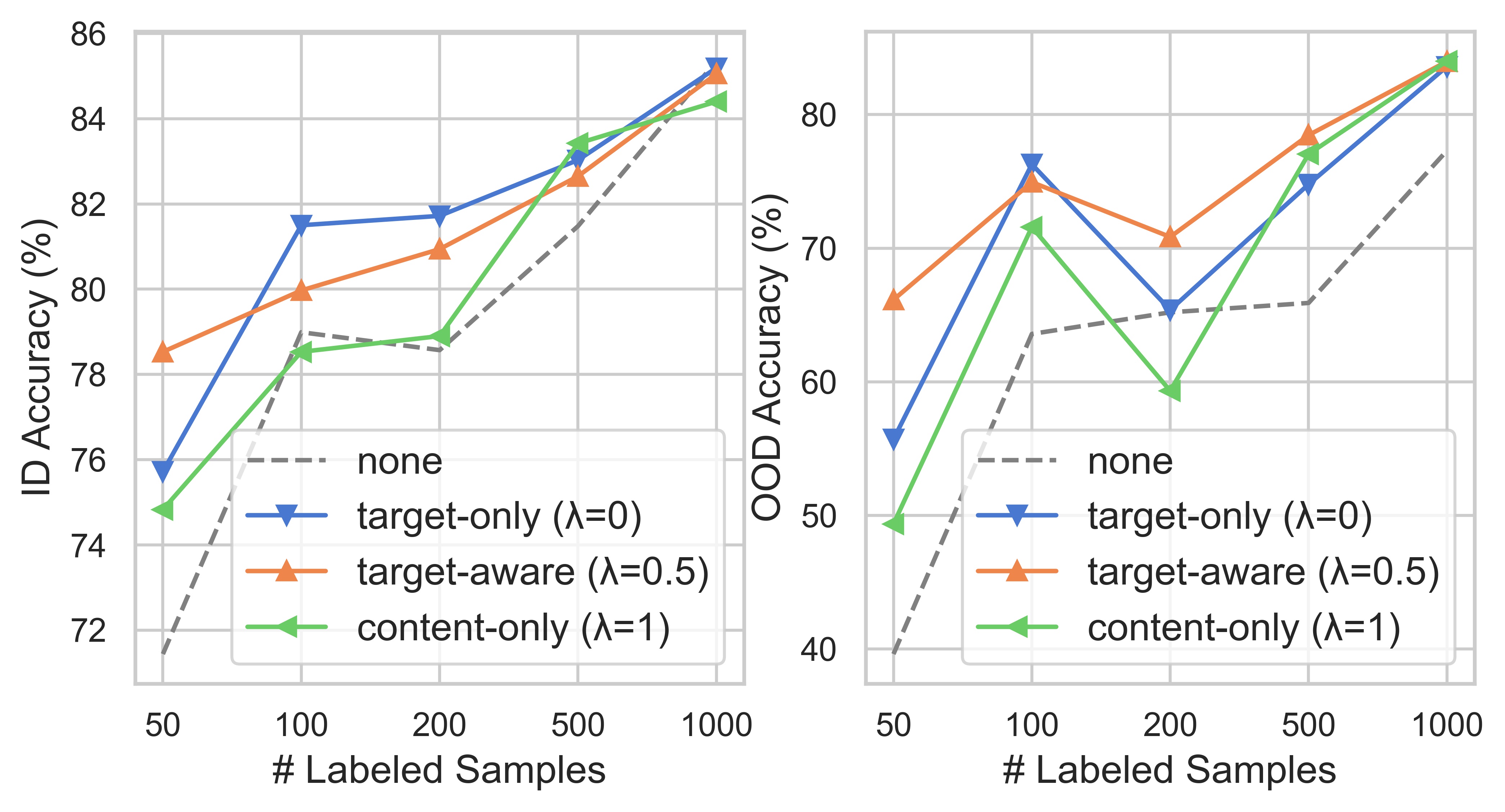}
\caption{\small{Comparison of different sketch extraction strategies on the Huff dataset with different labeled sizes.}}
\label{lambda-impact}
\end{figure}

\subsubsection{Ablation Study}
\textbf{\ours{} pre-training.}
\ours{} is pre-trained using a novel \textit{reconstruction from sketch} objective, which enables the model to reconstruct sentences or paragraphs given only a few segments. The backbone of \ours{} – BART \cite{lewis2020bart} can also be used to reconstruct text where certain short spans (ranging from 1-3 words) are masked. Therefore, we can verify the effectiveness of \ours{}'s pre-training by directly using BART-large's infilling ability for our proposed sketch-based generation or sketch-based fine-tuning, without the \ours{}-like pre-training, denoted as "$-$ PT". We experiment on the Huff dataset and Huff$\Rightarrow$BBC task as shown in Table \ref{ablation-pretrain}. 
Compared with \oursaug{} or \oursaug{}-f, using BART for sketch-based generation achieves significantly worse results, especially in the OOD or low-resource setting. The pre-training of BART determines it can only reconstruct small textual corruption. In comparison, \oursaug{} uses a much larger masking ratio during pre-training, resulting in the strong ability to generate complete and coherent context around the sketch. These results show the importance of our proposed \textit{reconstruction from sketch} pre-training.

\noindent\textbf{Target-aware sketch extraction.}
 Recall that our target-aware sketch extraction uses a fused embedding from both the document and the target label, by using a hyper-parameter $\lambda=0.5$ to balance the information from both sides. Now we compare the \textit{content-only} and \textit{task-only} strategies by setting $\lambda=1$ and $\lambda=0$ respectively. Figure \ref{lambda-impact} shows the comparison of these different strategies on the Huff dataset with different training sizes. Results reveal that target-aware is more robust than the other two strategies in both ID and OOD scenarios, while content-only is relatively the worst across all settings. Extracting the sketches based only on the content may lose some key information related to downstream tasks, while based only on the target may result in a lack of diversity. It is beneficial to consider both sides for sketch construction, which shows the effectiveness of our proposed target-aware sketch extraction. Note that $\lambda=0.5$ may not be optimal for every dataset therefore we encourage tuning this hyper-parameter according to the specific task.

\section{Related Work}\label{sec:related-work}
\subsection{Reconstruction for Corrupted Text}\label{sec:related-word-denoising}
Many recent pretrained transformer models (PTMs) are built on denoising pre-training, with text reconstruction as (one of) their pre-training objective(s). BERT \cite{devlin2018bert} first uses a masked language modeling (MLM) task in pre-training, which masks 15\% of the tokens in the original text and asks the model to predict the missing tokens. The MLM task and the 15\% masking ratio are also used by BERT's successors, like RoBERTa \cite{liu2019roberta} and ALBERT \cite{albert}. BART \cite{lewis2020bart} uses a novel text infilling task with a bigger masking ratio of 30\% for pre-training, where small spans of text are corrupted for reconstruction. MASS \cite{song2019mass} uses a higher masking ratio (50\%) for seq2seq pre-training, but is limited to sentence level with only one consecutive fragment being masked for each sentence, which makes MASS unable to reconstruct the text based on a sketch that may have several blanks. GLM \cite{du2022glm} pre-trains a self-attention Transformer using an autoregressive blank infilling objective, also with a 15\% random masking. However, GLM predicts the masked spans in an iterative manner where the current prediction should be based on previously predicted spans, limiting the efficiency of sketch-based text generation. To the best of our knowledge, \ours{} is the first language model pre-trained with \textit{extreme and selective masking} on the large-scale general corpus.

Our work is also related to text-infilling methods, including \citet{zhu2019text-infilling}'s, TIGS \cite{liu2019tigs}, blank language models \cite{shen2020blank}, ILM \cite{ilm}, and lexically constrained generation methods like POINTER \cite{zhang2020pointer} and CBART \cite{cbart}. Keywords-to-text methods like T5-CommonGen \cite{gehrmann2021gem} also shares some similarity. However, these methods are not designed for the sketch-based text generation task defined in this work. They are not suitable for this task for \textit{at least one} of the following reasons: 1) limited to unigrams as input; 2) mainly trained for short-sentence reconstruction; 3) fixed blank-filling length; 3) traditional model architecture without pre-training; 4) no pre-training on general corpus for public use. These limitations will result in high sketch-lost or in-fluency of the generated text. We compare with some of the typical methods in Section \ref{sec:nlg-eval}, which shows the superiority of our proposed \ours{}.

Recently \citet{zeng2022glm-130b} released a giant pre-trained model – GLM-130B, containing 130 billion parameters. GLM-130B also has strong abilities for blank-filling tasks. Due to the limited time and resources, we didn't compare with it in the experiments part. In Appendix \ref{sec:more-nlg-examples}, we show some examples generated by GLM-130B by calling their online API and analyze the advantages and disadvantages of both \ours{} and GLM-130B.

\subsection{Data Augmentation in NLP}
Data augmentation techniques are extensively studied in all kinds of NLP tasks. Most of these methods conduct augmentation on the input space by generating new training samples \cite{da_survey}, while some other methods are applied to the feature space, such as embedding mixup \cite{mixup_text,mixup_transformer,chen2020mixtext}. In this work, we mainly focus on the input space data augmentation. Rule-based methods are widely used for various tasks, including text classification tasks \cite{wei2019eda,guo2022sta}, named entity recognition \cite{simple-ner-aug} and natural language generation \cite{feng-etal-2020-genaug}. Apart from the rule-based methods, there are also plenty of generative model-based augmentation methods, such as back-translation \cite{back_translation,back_translation1}, paraphrasing \cite{gao2020paraphrase,2021parrot_paraphrase}, contextual augmentation \cite{cbert, PTM4aug} that utilize the masking mechanism in masked language models (MLM) and open-ended generation methods \cite{lambada, PTM4aug} that use auto-regressive (AR) models like GPT-2 \cite{GPT-2} to generate new text by fine-tuning the model to generate the original text conditioned on the label.

Our proposed \oursaug{} is also on the line of generative augmentation methods, but is different from previous generative methods in the following ways: Compared with methods like MLM-based or paraphrasing methods, \oursaug{} introduces more diversity to the training set; Compared with AR-based open generation methods, \oursaug{} is more controllable in the content and quality of generated samples; Methods like C-MLM \cite{cbert,PTM4aug}, LAMBADA \cite{lambada} and MELM \cite{zhou2022melm} also involve an extra fine-tuning step on the downstream datasets, making them more inconvenient for deployment, while \oursaug{} can be directly used for augmentation (though we also show fine-tuning can lead to further performance gains); Last but not least, \oursaug{} is a more general approach applicable to a variety of NLP tasks while most of the traditional methods are task-specific.

\section{Discussions \& Limitations}\label{sec:limitations}
\subsection{\ours{} Model.}
In this work, we present the sketch-based text generation task, a pre-trained model \ours{} specifically designed for this task, and a novel \oursaug{} method that effectively applies sketch-based text generation to data augmentation scenarios. Sketch-based text generation task might be a new task for previous methods, which partly explains why they are inferior to \ours{} in our experiments. Therefore, we don't expect \ours{} to be the state-of-the-art (SOTA) model for all conditional language generation tasks. 

Our current research on \ours{} also has some limitations, which are under our future research. For example, using more data and a larger backbone model can hopefully further improve the performance for more general use; the length of the generated text for each masked place cannot be controlled, limiting the usage for precise controlling; more experiments and analysis of using different pre-training strategies (such as the strategy used in ILM \cite{ilm}) should be studied.

\subsection{\oursaug{}.}
Though \oursaug{} shows advantages over a bunch of previous augmentation methods in a variety of NLP tasks, we don't pursue making \oursaug{} the go-to augmentation tool for each specific task. The following limitations should be considered when using \oursaug{} for data augmentation:\\
1) Compared with rule-based methods, \oursaug{} needs more computation resources since it is based on a large pre-trained language model. We also release lighter versions trained on BART-base, but we haven't evaluated them on augmentation tasks; 2) \oursaug{}'s sketch extraction process is less compatible with tasks that require logical reasoning, such as natural language inference (NLI). This is because the extracted sketches may not be a good representation of the logical relationships contained in the text, which may result in noisy samples. For these tasks, we recommend utilizing the in-context learning of GPT-3 \cite{GPT-3}, or more fine-grained methods like FlipDA \cite{zhou2022flipda}; 3) When using \oursaug{} for NER data augmentation, new entities may be generated, which may lead to the unlabeled entity problem. To tackle this issue, an extra filtering step should be involved or utilize a modified loss function as described in Section \ref{sec:appendix-preprocessing}.

Therefore, we encourage combining different augmentation methods according to the characteristics of the downstream tasks.

\section{Conclusion}
In this paper, we present a sketch-based text generative model called \ours{}, which is pre-trained on a large-scale corpus with a novel extreme-and-selective masking strategy. Based on \ours{}, we propose a novel textual data augmentation method named \oursaug{} which can generate diverse and high-quality texts given textual sketches extracted from the training set. \oursaug{} can be directly applied to various NLP tasks without further fine-tuning on downstream datasets. Extensive experiments reveal the strong performance of \oursaug{} on classification, NER, and MRC tasks. In future work, we will further explore other applications of \ours{} and study the potential of using extreme-and-selective masking strategy to further improve current pre-trained language models.


\newpage
\bibliography{anthology,custom}

\begin{thebibliography}{66}
\expandafter\ifx\csname natexlab\endcsname\relax\def\natexlab#1{#1}\fi

\bibitem[{Alberti et~al.(2019)Alberti, Andor, Pitler, Devlin, and
  Collins}]{alberti2019synthetic}
Chris Alberti, Daniel Andor, Emily Pitler, Jacob Devlin, and Michael Collins.
  2019.
\newblock Synthetic qa corpora generation with roundtrip consistency.
\newblock In \emph{Proceedings of the 57th Annual Meeting of the Association
  for Computational Linguistics}, pages 6168--6173.

\bibitem[{Anaby-Tavor et~al.(2020)Anaby-Tavor, Carmeli, Goldbraich, Kantor,
  Kour, Shlomov, Tepper, and Zwerdling}]{lambada}
Ateret Anaby-Tavor, Boaz Carmeli, Esther Goldbraich, Amir Kantor, George Kour,
  Segev Shlomov, Naama Tepper, and Naama Zwerdling. 2020.
\newblock Do not have enough data? deep learning to the rescue!
\newblock In \emph{Proceedings of the AAAI Conference on Artificial
  Intelligence}, volume~34, pages 7383--7390.

\bibitem[{Barratt and Sharma(2018)}]{barratt2018note-IS}
Shane Barratt and Rishi Sharma. 2018.
\newblock A note on the inception score.
\newblock \emph{arXiv preprint arXiv:1801.01973}.

\bibitem[{Brown et~al.(2020)Brown, Mann, Ryder, Subbiah, Kaplan, Dhariwal,
  Neelakantan, Shyam, Sastry, Askell et~al.}]{GPT-3}
Tom Brown, Benjamin Mann, Nick Ryder, Melanie Subbiah, Jared~D Kaplan, Prafulla
  Dhariwal, Arvind Neelakantan, Pranav Shyam, Girish Sastry, Amanda Askell,
  et~al. 2020.
\newblock Language models are few-shot learners.
\newblock \emph{Advances in neural information processing systems},
  33:1877--1901.

\bibitem[{Campos et~al.(2020)Campos, Mangaravite, Pasquali, Jorge, Nunes, and
  Jatowt}]{campos2020yake}
Ricardo Campos, V{\'\i}tor Mangaravite, Arian Pasquali, Al{\'\i}pio Jorge,
  C{\'e}lia Nunes, and Adam Jatowt. 2020.
\newblock Yake! keyword extraction from single documents using multiple local
  features.
\newblock \emph{Information Sciences}, 509:257--289.

\bibitem[{Chen et~al.(2020)Chen, Yang, and Yang}]{chen2020mixtext}
Jiaao Chen, Zichao Yang, and Diyi Yang. 2020.
\newblock Mixtext: Linguistically-informed interpolation of hidden space for
  semi-supervised text classification.
\newblock In \emph{Proceedings of the 58th Annual Meeting of the Association
  for Computational Linguistics}, pages 2147--2157.

\bibitem[{Dai and Adel(2020)}]{simple-ner-aug}
Xiang Dai and Heike Adel. 2020.
\newblock An analysis of simple data augmentation for named entity recognition.
\newblock In \emph{Proceedings of the 28th International Conference on
  Computational Linguistics}, Barcelona, Spain (Online). International
  Committee on Computational Linguistics.

\bibitem[{Damodaran(2021)}]{2021parrot_paraphrase}
Prithiviraj Damodaran. 2021.
\newblock Parrot: Paraphrase generation for nlu.

\bibitem[{Dathathri et~al.(2019)Dathathri, Madotto, Lan, Hung, Frank, Molino,
  Yosinski, and Liu}]{pplm}
Sumanth Dathathri, Andrea Madotto, Janice Lan, Jane Hung, Eric Frank, Piero
  Molino, Jason Yosinski, and Rosanne Liu. 2019.
\newblock Plug and play language models: A simple approach to controlled text
  generation.
\newblock In \emph{International Conference on Learning Representations}.

\bibitem[{Devlin et~al.(2019)Devlin, Chang, Lee, and
  Toutanova}]{devlin2018bert}
Jacob Devlin, Ming-Wei Chang, Kenton Lee, and Kristina Toutanova. 2019.
\newblock Bert: Pre-training of deep bidirectional transformers for language
  understanding.
\newblock In \emph{NAACL-HLT (1)}.

\bibitem[{Donahue et~al.(2020)Donahue, Lee, and Liang}]{ilm}
Chris Donahue, Mina Lee, and Percy Liang. 2020.
\newblock Enabling language models to fill in the blanks.
\newblock In \emph{Proceedings of the 58th Annual Meeting of the Association
  for Computational Linguistics}, pages 2492--2501.

\bibitem[{Du et~al.(2022)Du, Qian, Liu, Ding, Qiu, Yang, and Tang}]{du2022glm}
Zhengxiao Du, Yujie Qian, Xiao Liu, Ming Ding, Jiezhong Qiu, Zhilin Yang, and
  Jie Tang. 2022.
\newblock Glm: General language model pretraining with autoregressive blank
  infilling.
\newblock In \emph{Proceedings of the 60th Annual Meeting of the Association
  for Computational Linguistics (Volume 1: Long Papers)}, pages 320--335.

\bibitem[{Feng et~al.(2020)Feng, Gangal, Kang, Mitamura, and
  Hovy}]{feng-etal-2020-genaug}
Steven~Y. Feng, Varun Gangal, Dongyeop Kang, Teruko Mitamura, and Eduard Hovy.
  2020.
\newblock \href {https://doi.org/10.18653/v1/2020.deelio-1.4} {{G}en{A}ug: Data
  augmentation for finetuning text generators}.
\newblock In \emph{Proceedings of Deep Learning Inside Out (DeeLIO): The First
  Workshop on Knowledge Extraction and Integration for Deep Learning
  Architectures}, pages 29--42, Online. Association for Computational
  Linguistics.

\bibitem[{Feng et~al.(2021)Feng, Gangal, Wei, Chandar, Vosoughi, Mitamura, and
  Hovy}]{da_survey}
Steven~Y Feng, Varun Gangal, Jason Wei, Sarath Chandar, Soroush Vosoughi,
  Teruko Mitamura, and Eduard Hovy. 2021.
\newblock A survey of data augmentation approaches for nlp.
\newblock In \emph{Findings of the Association for Computational Linguistics:
  ACL-IJCNLP 2021}, pages 968--988.

\bibitem[{French(1999)}]{french1999catastrophic}
Robert~M French. 1999.
\newblock Catastrophic forgetting in connectionist networks.
\newblock \emph{Trends in cognitive sciences}, 3(4):128--135.

\bibitem[{Gao et~al.(2020)Gao, Zhang, Ou, and Yu}]{gao2020paraphrase}
Silin Gao, Yichi Zhang, Zhijian Ou, and Zhou Yu. 2020.
\newblock Paraphrase augmented task-oriented dialog generation.
\newblock In \emph{Proceedings of the 58th Annual Meeting of the Association
  for Computational Linguistics}, pages 639--649.

\bibitem[{Gehrmann et~al.(2021)Gehrmann, Adewumi, Aggarwal, Ammanamanchi,
  Aremu, Bosselut, Chandu, Clinciu, Das, Dhole et~al.}]{gehrmann2021gem}
Sebastian Gehrmann, Tosin Adewumi, Karmanya Aggarwal, Pawan~Sasanka
  Ammanamanchi, Anuoluwapo Aremu, Antoine Bosselut, Khyathi~Raghavi Chandu,
  Miruna-Adriana Clinciu, Dipanjan Das, Kaustubh Dhole, et~al. 2021.
\newblock The gem benchmark: Natural language generation, its evaluation and
  metrics.
\newblock In \emph{Proceedings of the 1st Workshop on Natural Language
  Generation, Evaluation, and Metrics (GEM 2021)}, pages 96--120.

\bibitem[{Greene and Cunningham(2006)}]{data-bbc}
Derek Greene and P\'{a}draig Cunningham. 2006.
\newblock Practical solutions to the problem of diagonal dominance in kernel
  document clustering.
\newblock In \emph{Proc. 23rd International Conference on Machine learning
  (ICML'06)}, pages 377--384. ACM Press.

\bibitem[{Guo et~al.(2022)Guo, Han, and Huang}]{guo2022sta}
Biyang Guo, Songqiao Han, and Hailiang Huang. 2022.
\newblock Selective text augmentation with word roles for low-resource text
  classification.
\newblock \emph{arXiv preprint arXiv:2209.01560}.

\bibitem[{Guo et~al.(2019)Guo, Mao, and Zhang}]{mixup_text}
Hongyu Guo, Yongyi Mao, and Richong Zhang. 2019.
\newblock Augmenting data with mixup for sentence classification: An empirical
  study.
\newblock \emph{arXiv preprint arXiv:1905.08941}.

\bibitem[{He(2021)}]{cbart}
Xingwei He. 2021.
\newblock Parallel refinements for lexically constrained text generation with
  bart.
\newblock In \emph{Proceedings of the 2021 Conference on Empirical Methods in
  Natural Language Processing}, pages 8653--8666.

\bibitem[{Hendrycks et~al.(2020)Hendrycks, Liu, Wallace, Dziedzic, Krishnan,
  and Song}]{PTM-OOD}
Dan Hendrycks, Xiaoyuan Liu, Eric Wallace, Adam Dziedzic, Rishabh Krishnan, and
  Dawn Song. 2020.
\newblock Pretrained transformers improve out-of-distribution robustness.
\newblock In \emph{Proceedings of the 58th Annual Meeting of the Association
  for Computational Linguistics}, pages 2744--2751.

\bibitem[{Hochreiter and Schmidhuber(1997)}]{hochreiter1997lstm}
Sepp Hochreiter and J{\"u}rgen Schmidhuber. 1997.
\newblock Long short-term memory.
\newblock \emph{Neural computation}, 9(8):1735--1780.

\bibitem[{Jelinek et~al.(1977)Jelinek, Mercer, Bahl, and
  Baker}]{jelinek1977perplexity}
Fred Jelinek, Robert~L Mercer, Lalit~R Bahl, and James~K Baker. 1977.
\newblock Perplexity—a measure of the difficulty of speech recognition tasks.
\newblock \emph{The Journal of the Acoustical Society of America},
  62(S1):S63--S63.

\bibitem[{Keskar et~al.(2019)Keskar, McCann, Varshney, Xiong, and
  Socher}]{keskar2019ctrl}
Nitish~Shirish Keskar, Bryan McCann, Lav~R Varshney, Caiming Xiong, and Richard
  Socher. 2019.
\newblock Ctrl: A conditional transformer language model for controllable
  generation.
\newblock \emph{arXiv preprint arXiv:1909.05858}.

\bibitem[{Kumar et~al.(2020)Kumar, Choudhary, and Cho}]{PTM4aug}
Varun Kumar, Ashutosh Choudhary, and Eunah Cho. 2020.
\newblock Data augmentation using pre-trained transformer models.
\newblock In \emph{Proceedings of the 2nd Workshop on Life-long Learning for
  Spoken Language Systems}, pages 18--26.

\bibitem[{Lan et~al.(2019)Lan, Chen, Goodman, Gimpel, Sharma, and
  Soricut}]{albert}
Zhenzhong Lan, Mingda Chen, Sebastian Goodman, Kevin Gimpel, Piyush Sharma, and
  Radu Soricut. 2019.
\newblock Albert: A lite bert for self-supervised learning of language
  representations.
\newblock \emph{arXiv preprint arXiv:1909.11942}.

\bibitem[{Lewis et~al.(2020)Lewis, Liu, Goyal, Ghazvininejad, Mohamed, Levy,
  Stoyanov, and Zettlemoyer}]{lewis2020bart}
Mike Lewis, Yinhan Liu, Naman Goyal, Marjan Ghazvininejad, Abdelrahman Mohamed,
  Omer Levy, Veselin Stoyanov, and Luke Zettlemoyer. 2020.
\newblock Bart: Denoising sequence-to-sequence pre-training for natural
  language generation, translation, and comprehension.
\newblock In \emph{Proceedings of the 58th Annual Meeting of the Association
  for Computational Linguistics}, pages 7871--7880.

\bibitem[{Li et~al.(2020)Li, Shi et~al.}]{iclr-ner-negative}
Yangming Li, Shuming Shi, et~al. 2020.
\newblock Empirical analysis of unlabeled entity problem in named entity
  recognition.
\newblock In \emph{International Conference on Learning Representations}.

\bibitem[{Lin et~al.(2020)Lin, Zhou, Shen, Zhou, Bhagavatula, Choi, and
  Ren}]{lin2020commongen}
Bill~Yuchen Lin, Wangchunshu Zhou, Ming Shen, Pei Zhou, Chandra Bhagavatula,
  Yejin Choi, and Xiang Ren. 2020.
\newblock Commongen: A constrained text generation challenge for generative
  commonsense reasoning.
\newblock In \emph{Findings of the Association for Computational Linguistics:
  EMNLP 2020}, pages 1823--1840.

\bibitem[{Liu et~al.(2019{\natexlab{a}})Liu, Fu, Liu, and Lv}]{liu2019tigs}
Dayiheng Liu, Jie Fu, Pengfei Liu, and Jiancheng Lv. 2019{\natexlab{a}}.
\newblock Tigs: An inference algorithm for text infilling with gradient search.
\newblock In \emph{Proceedings of the 57th Annual Meeting of the Association
  for Computational Linguistics}, pages 4146--4156.

\bibitem[{Liu et~al.(2020)Liu, Gong, Fu, Yan, Chen, Lv, Duan, and Zhou}]{CRQDA}
Dayiheng Liu, Yeyun Gong, Jie Fu, Yu~Yan, Jiusheng Chen, Jiancheng Lv, Nan
  Duan, and Ming Zhou. 2020.
\newblock Tell me how to ask again: Question data augmentation with
  controllable rewriting in continuous space.
\newblock In \emph{Proceedings of the 2020 Conference on Empirical Methods in
  Natural Language Processing (EMNLP)}, pages 5798--5810.

\bibitem[{Liu et~al.(2019{\natexlab{b}})Liu, Ott, Goyal, Du, Joshi, Chen, Levy,
  Lewis, Zettlemoyer, and Stoyanov}]{liu2019roberta}
Yinhan Liu, Myle Ott, Naman Goyal, Jingfei Du, Mandar Joshi, Danqi Chen, Omer
  Levy, Mike Lewis, Luke Zettlemoyer, and Veselin Stoyanov. 2019{\natexlab{b}}.
\newblock Roberta: A robustly optimized bert pretraining approach.
\newblock \emph{arXiv preprint arXiv:1907.11692}.

\bibitem[{Loshchilov and Hutter(2017)}]{adamw}
Ilya Loshchilov and Frank Hutter. 2017.
\newblock Decoupled weight decay regularization.
\newblock \emph{arXiv preprint arXiv:1711.05101}.

\bibitem[{Maas et~al.(2011)Maas, Daly, Pham, Huang, Ng, and Potts}]{data-imdb}
Andrew~L. Maas, Raymond~E. Daly, Peter~T. Pham, Dan Huang, Andrew~Y. Ng, and
  Christopher Potts. 2011.
\newblock \href {http://www.aclweb.org/anthology/P11-1015} {Learning word
  vectors for sentiment analysis}.
\newblock In \emph{Proceedings of the 49th Annual Meeting of the Association
  for Computational Linguistics: Human Language Technologies}, pages 142--150,
  Portland, Oregon, USA. Association for Computational Linguistics.

\bibitem[{Misra and Grover(2021)}]{huffpost}
Rishabh Misra and Jigyasa Grover. 2021.
\newblock \emph{Sculpting Data for ML: The first act of Machine Learning}.

\bibitem[{Morris et~al.(2004)Morris, Maier, and Green}]{morris2004CER}
Andrew Morris, Viktoria Maier, and Phil Green. 2004.
\newblock From wer and ril to mer and wil: improved evaluation measures for
  connected speech recognition.

\bibitem[{Papineni et~al.(2002)Papineni, Roukos, Ward, and jing
  Zhu}]{Papineni02bleu:a}
Kishore Papineni, Salim Roukos, Todd Ward, and Wei jing Zhu. 2002.
\newblock Bleu: a method for automatic evaluation of machine translation.
\newblock pages 311--318.

\bibitem[{Radford et~al.(2019)Radford, Wu, Child, Luan, Amodei, Sutskever
  et~al.}]{GPT-2}
Alec Radford, Jeffrey Wu, Rewon Child, David Luan, Dario Amodei, Ilya
  Sutskever, et~al. 2019.
\newblock Language models are unsupervised multitask learners.
\newblock \emph{OpenAI blog}, 1(8):9.

\bibitem[{Raffel et~al.(2019)Raffel, Shazeer, Roberts, Lee, Narang, Matena,
  Zhou, Li, and Liu}]{2019t5-c4}
Colin Raffel, Noam Shazeer, Adam Roberts, Katherine Lee, Sharan Narang, Michael
  Matena, Yanqi Zhou, Wei Li, and Peter~J. Liu. 2019.
\newblock \href {http://arxiv.org/abs/1910.10683} {Exploring the limits of
  transfer learning with a unified text-to-text transformer}.
\newblock \emph{arXiv e-prints}.

\bibitem[{Rajpurkar et~al.(2016)Rajpurkar, Zhang, Lopyrev, and
  Liang}]{data-squad}
Pranav Rajpurkar, Jian Zhang, Konstantin Lopyrev, and Percy Liang. 2016.
\newblock Squad: 100,000+ questions for machine comprehension of text.
\newblock \emph{arXiv preprint arXiv:1606.05250}.

\bibitem[{Reimers and Gurevych(2019)}]{sentence-bert}
Nils Reimers and Iryna Gurevych. 2019.
\newblock Sentence-bert: Sentence embeddings using siamese bert-networks.
\newblock In \emph{Proceedings of the 2019 Conference on Empirical Methods in
  Natural Language Processing and the 9th International Joint Conference on
  Natural Language Processing (EMNLP-IJCNLP)}, pages 3982--3992.

\bibitem[{Sang and Meulder(2003)}]{data-conll03}
Erik Tjong~Kim Sang and Fien~De Meulder. 2003.
\newblock Introduction to the conll-2003 shared task: language-independent
  named entity recognition.
\newblock \emph{conference on computational natural language learning}.

\bibitem[{Sanh et~al.(2019)Sanh, Debut, Chaumond, and
  Wolf}]{sanh2019distilbert}
Victor Sanh, Lysandre Debut, Julien Chaumond, and Thomas Wolf. 2019.
\newblock Distilbert, a distilled version of bert: smaller, faster, cheaper and
  lighter.
\newblock \emph{arXiv preprint arXiv:1910.01108}.

\bibitem[{Sennrich et~al.(2016)Sennrich, Haddow, and Birch}]{back_translation}
Rico Sennrich, Barry Haddow, and Alexandra Birch. 2016.
\newblock Improving neural machine translation models with monolingual data.
\newblock In \emph{Proceedings of the 54th Annual Meeting of the Association
  for Computational Linguistics (Volume 1: Long Papers)}, pages 86--96.

\bibitem[{Shao et~al.(2021)Shao, Geng, Liu, Dai, Yang, Zhe, Bao, and
  Qiu}]{shao2021cpt-bart-chinese}
Yunfan Shao, Zhichao Geng, Yitao Liu, Junqi Dai, Fei Yang, Li~Zhe, Hujun Bao,
  and Xipeng Qiu. 2021.
\newblock Cpt: A pre-trained unbalanced transformer for both chinese language
  understanding and generation.
\newblock \emph{arXiv preprint arXiv:2109.05729}.

\bibitem[{Shen et~al.(2020)Shen, Quach, Barzilay, and Jaakkola}]{shen2020blank}
Tianxiao Shen, Victor Quach, Regina Barzilay, and Tommi Jaakkola. 2020.
\newblock Blank language models.
\newblock In \emph{Proceedings of the 2020 Conference on Empirical Methods in
  Natural Language Processing (EMNLP)}, pages 5186--5198.

\bibitem[{Shih et~al.(2019)Shih, Chang, and Yang}]{Shih2019XLEditorPS}
Yong-Siang Shih, Wei-Cheng Chang, and Yiming Yang. 2019.
\newblock Xl-editor: Post-editing sentences with xlnet.
\newblock \emph{ArXiv}, abs/1910.10479.

\bibitem[{Silfverberg et~al.(2017)Silfverberg, Wiemerslage, Liu, and
  Mao}]{back_translation2}
Miikka Silfverberg, Adam Wiemerslage, Ling Liu, and Lingshuang~Jack Mao. 2017.
\newblock Data augmentation for morphological reinflection.
\newblock In \emph{Proceedings of the CoNLL SIGMORPHON 2017 Shared Task:
  Universal Morphological Reinflection}, pages 90--99.

\bibitem[{Socher et~al.(2013)Socher, Perelygin, Wu, Chuang, Manning, Ng, and
  Potts}]{Socher-sst2}
Richard Socher, Alex Perelygin, Jean Wu, Jason Chuang, Christopher~D Manning,
  Andrew~Y Ng, and Christopher Potts. 2013.
\newblock Recursive deep models for semantic compositionality over a sentiment
  treebank.
\newblock In \emph{Proceedings of the 2013 conference on empirical methods in
  natural language processing}, pages 1631--1642.

\bibitem[{Song et~al.(2019)Song, Tan, Qin, Lu, and Liu}]{song2019mass}
Kaitao Song, Xu~Tan, Tao Qin, Jianfeng Lu, and Tie-Yan Liu. 2019.
\newblock Mass: Masked sequence to sequence pre-training for language
  generation.
\newblock In \emph{International Conference on Machine Learning}, pages
  5926--5936. PMLR.

\bibitem[{Sun et~al.(2020)Sun, Xia, Yin, Liang, Yu, and He}]{mixup_transformer}
Lichao Sun, Congying Xia, Wenpeng Yin, Tingting Liang, Philip~S Yu, and Lifang
  He. 2020.
\newblock Mixup-transformer: Dynamic data augmentation for nlp tasks.
\newblock \emph{arXiv preprint arXiv:2010.02394}.

\bibitem[{Tiedemann(2020)}]{tiedemann-2020-tatoeba-translation-model}
J{\"o}rg Tiedemann. 2020.
\newblock \href {https://www.aclweb.org/anthology/2020.wmt-1.139} {The
  {T}atoeba {T}ranslation {C}hallenge {--} {R}ealistic data sets for low
  resource and multilingual {MT}}.
\newblock In \emph{Proceedings of the Fifth Conference on Machine Translation},
  pages 1174--1182, Online. Association for Computational Linguistics.

\bibitem[{Wei and Zou(2019)}]{wei2019eda}
Jason Wei and Kai Zou. 2019.
\newblock \href {https://doi.org/10.18653/v1/D19-1670} {{EDA}: Easy data
  augmentation techniques for boosting performance on text classification
  tasks}.
\newblock In \emph{Proceedings of the 2019 Conference on Empirical Methods in
  Natural Language Processing and the 9th International Joint Conference on
  Natural Language Processing (EMNLP-IJCNLP)}, Hong Kong, China. Association
  for Computational Linguistics.

\bibitem[{Wu et~al.(2019)Wu, Lv, Zang, Han, and Hu}]{cbert}
Xing Wu, Shangwen Lv, Liangjun Zang, Jizhong Han, and Songlin Hu. 2019.
\newblock Conditional bert contextual augmentation.
\newblock In \emph{International Conference on Computational Science}, pages
  84--95. Springer.

\bibitem[{Xu(2019)}]{clue-corpus}
Bright Xu. 2019.
\newblock \href {https://doi.org/10.5281/zenodo.3402023} {Nlp chinese corpus:
  Large scale chinese corpus for nlp}.

\bibitem[{Yao et~al.(2019)Yao, Peng, Weischedel, Knight, Zhao, and
  Yan}]{yao2019plan-and-write}
Lili Yao, Nanyun Peng, Ralph Weischedel, Kevin Knight, Dongyan Zhao, and Rui
  Yan. 2019.
\newblock Plan-and-write: Towards better automatic storytelling.
\newblock In \emph{Proceedings of the AAAI Conference on Artificial
  Intelligence}, volume~33, pages 7378--7385.

\bibitem[{Yu et~al.(2018)Yu, Dohan, Luong, Zhao, Chen, Norouzi, and
  Le}]{back_translation1}
Adams~Wei Yu, David Dohan, Minh-Thang Luong, Rui Zhao, Kai Chen, Mohammad
  Norouzi, and Quoc~V Le. 2018.
\newblock Qanet: Combining local convolution with global self-attention for
  reading comprehension.
\newblock \emph{arXiv preprint arXiv:1804.09541}.

\bibitem[{Zeng et~al.(2022)Zeng, Liu, Du, Wang, Lai, Ding, Yang, Xu, Zheng, Xia
  et~al.}]{zeng2022glm-130b}
Aohan Zeng, Xiao Liu, Zhengxiao Du, Zihan Wang, Hanyu Lai, Ming Ding, Zhuoyi
  Yang, Yifan Xu, Wendi Zheng, Xiao Xia, et~al. 2022.
\newblock Glm-130b: An open bilingual pre-trained model.
\newblock \emph{arXiv preprint arXiv:2210.02414}.

\bibitem[{Zhang et~al.(2018)Zhang, Cisse, Dauphin, and
  Lopez-Paz}]{zhang2018mixup}
Hongyi Zhang, Moustapha Cisse, Yann~N Dauphin, and David Lopez-Paz. 2018.
\newblock mixup: Beyond empirical risk minimization.
\newblock In \emph{International Conference on Learning Representations}.

\bibitem[{Zhang et~al.(2015)Zhang, Zhao, and LeCun}]{data-zhang-cnn}
Xiang Zhang, Junbo Zhao, and Yann LeCun. 2015.
\newblock Character-level convolutional networks for text classification.
\newblock \emph{Advances in neural information processing systems}, 28.

\bibitem[{Zhang et~al.(2020)Zhang, Wang, Li, Gan, Brockett, and
  Dolan}]{zhang2020pointer}
Yizhe Zhang, Guoyin Wang, Chunyuan Li, Zhe Gan, Chris Brockett, and William~B
  Dolan. 2020.
\newblock Pointer: Constrained progressive text generation via insertion-based
  generative pre-training.
\newblock In \emph{Proceedings of the 2020 Conference on Empirical Methods in
  Natural Language Processing (EMNLP)}, pages 8649--8670.

\bibitem[{Zhou et~al.(2022{\natexlab{a}})Zhou, Zheng, Tang, Jian, and
  Yang}]{zhou2022flipda}
Jing Zhou, Yanan Zheng, Jie Tang, Li~Jian, and Zhilin Yang. 2022{\natexlab{a}}.
\newblock Flipda: Effective and robust data augmentation for few-shot learning.
\newblock In \emph{Proceedings of the 60th Annual Meeting of the Association
  for Computational Linguistics (Volume 1: Long Papers)}, pages 8646--8665.

\bibitem[{Zhou et~al.(2022{\natexlab{b}})Zhou, Li, He, Bing, Cambria, Si, and
  Miao}]{zhou2022melm}
Ran Zhou, Xin Li, Ruidan He, Lidong Bing, Erik Cambria, Luo Si, and Chunyan
  Miao. 2022{\natexlab{b}}.
\newblock Melm: Data augmentation with masked entity language modeling for
  low-resource ner.
\newblock In \emph{Proceedings of the 60th Annual Meeting of the Association
  for Computational Linguistics (Volume 1: Long Papers)}, pages 2251--2262.

\bibitem[{Zhu et~al.(2019)Zhu, Hu, and Xing}]{zhu2019text-infilling}
Wanrong Zhu, Zhiting Hu, and Eric Xing. 2019.
\newblock Text infilling.
\newblock \emph{arXiv preprint arXiv:1901.00158}.

\bibitem[{Ziegler et~al.(2019)Ziegler, Stiennon, Wu, Brown, Radford, Amodei,
  Christiano, and Irving}]{gpt-2-finetune}
Daniel~M Ziegler, Nisan Stiennon, Jeffrey Wu, Tom~B Brown, Alec Radford, Dario
  Amodei, Paul Christiano, and Geoffrey Irving. 2019.
\newblock Fine-tuning language models from human preferences.
\newblock \emph{arXiv preprint arXiv:1909.08593}.

\end{thebibliography}
\bibliographystyle{acl_natbib}

\newpage
\appendix

\section{Experiment Details}\label{sec:appendix-exp-details}

\subsection{\ours{} Pre-training}\label{sec:pre-training}

To extract the sketch of a given document, we utilize YAKE \cite{campos2020yake}, a lightweight unsupervised automatic keyword extraction method to select the most relevant keywords. We set \texttt{max\_ngram=3} and \texttt{topk=max(l/5,10)} where \texttt{l} is the length of the document (number of words). Then a projection and a masking process are applied to obtain the sketch of the document. We sampled 27 million paragraphs (with less than 15 sentences in each paragraph for the base model and length ranging from 50 to 200 for the large model) as the training set from the \texttt{realnewslike} split of C4 dataset \cite{2019t5-c4}. Note that \texttt{topk=l/5} doesn't mean the masking ratio is 80\% (4/5), since the keywords may occur multiple times in the document or be contained within other keywords. We calculated the masking ratio of 1 million
<sketch,text> pairs randomly sampled from our training set, the average proportion (\%) is 72.97 ± 7.05.
 
\ours{} is a seq2seq model, with a bidirectional encoder and an auto-regressive decoder. \ours{} uses the same structure of BART \cite{lewis2020bart} and is initialized with the weights of BART-base or BART-large for \ours{}-base and \ours{}-large respectively. \ours{} is optimized using AdamW \cite{adamw} optimizer with learning rate 5.6e-5 and weight decay 0.01. We pre-train the \ours{} model for 3 epochs with batch size 32 using 8 NVIDIA V100 cards which takes a few days for the base model and around a week for the large model. 

The Chinese version \ours{}\texttt-base-chinese is trained on BART-base-chinese \cite{shao2021cpt-bart-chinese}, using 10 million passages from the CLUE corpus \cite{clue-corpus} for pre-training. A multilingual version is also under development.

All the code and models will be publicly available at \url{https://github.com/beyondguo/genius} and \url{https://github.com/microsoft/SCGLab}.

\subsection{Datasets \& Model Settings}\label{sec:appendix-exp-model}
For text classification, the training and validation sets are randomly sampled from the original datasets with $n \in \{50,100,200,500,1000\}$. We use the AdamW \cite{adamw} optimizer with learning rate 5e-5 for training and use early-stopping with patience=10 to choose the best model. We run all experiments with 5 random seeds and report the average performance.

For named entity recognition, we use the classical CoNLL03 dataset \cite{data-conll03} for evaluation. Notice that the consecutive sequences of the CoNLL03 dataset may come from the same article, thus may share some common entities. Previous works usually randomly sample $n$ sequences from the dataset, which may come from up to $n$ different articles, resulting in a sampled dataset with plenty of diverse entities. We claim that this approach does not correspond to a true low-resource scenario for NER tasks. Therefore, in this work, instead of randomly choosing $n$ samples from the dataset, we use the \textit{first-}$n$ samples from the dataset to construct training and validation sets, which are consecutive sequences coming from $m$ articles ($m << n$). We experiment on $n \in \{50,100,200,500\}$ for this task. We use BERT-base\cite{devlin2018bert} as the NER model, use AdamW optimizer with learning rate 2e-5 and linear learning rate scheduling for training. For each augmentation method, we train the model for 40 epochs and choose the best model according to the F1 score on the validation set.

For machine reading comprehension, we use the widely used SQuAD \cite{data-squad} dataset. The test set of SQuAD is not publicly available. To evaluate the performance in the test set, we need to corporate with the authors of SQuAD leaderboard. Since we mainly focus on the augmentation for low-resource setting in this paper where only a very small fraction of training data is used, we choose to only evaluate our method on the public development set, and use the same set of hyper-parameters for fair comparison among all the baselines. We use BERT-base as the basic model and the scripts provided by Huggingface\footnote{https://github.com/huggingface/transformers/ tree/main/examples/pytorch/question-answering} for model training and evaluating.

\subsection{Data Pre-Processing for Different NLP Tasks}\label{sec:appendix-preprocessing}
\textbf{Text classification.}\ \ 
 For text classification tasks, we use the categories of the samples as the TRI for target-aware sketches extraction during the augmentation of \ours{}. We use the \textbf{a}ttribute-\textbf{c}ontrolling (see Section \ref{sec:sega-augmentation}) to make the generated samples closer to their corresponding labels. Specifically, we add the label text as the prompt before the sketch, joined by \texttt{":"} or \texttt{"</s>"} in between.
 
 \noindent\textbf{NER.}\ \ 
 For NER, we use the entities in the training sequences as the TRI for target-aware sketch extraction, by which the extracted key-phrases are usually the textual spans that contain or semantically similar to the entities. The samples from the CoNLL03 dataset are usually short sentences, which are much shorter than the paragraphs used for \ours{} pre-training. Therefore, we concatenate the consecutive sequences into longer text before sketch extraction to make the inputs more compatible with \ours{}. Since \ours{} is a generative model which have potential to generate new entities that don't exist in the training set, resulting in unlabeled entity problem. To address this issue, we borrow the approach in \cite{iclr-ner-negative} by only labeling the entities that occur in the training set, leaving other tokens labeled with "X" (which don't contribute to the loss function). 
 
\noindent\textbf{MRC.}\ \ 
For MRC task, we use the widely used SQuAD dataset for experiment. Each training example of SQuAD is a triple of $(p, q, a)$ where $p$ is a multi-sentence paragraph that contains the answer $a$ to the question $q$. We use \ours{} to generate new paragraphs while keeping the question $q$ unchanged. To make the original answer $a$ accessible and reasonable for the generated paragraphs, we also keep the sentence where the answer occurs (noted as $s_{a}$) nearly unchanged and only augment the preceding ($p_{pre}$) and following text ($p_{post}$) of $s_a$. We use the current question as the TRI to extract sketches from $p_{pre}$ and $p_{post}$, noted as $s_{pre}$ and $s_{post}$ respectively. Then the sketch to the \ours{} model is the concatenation of $[s_{pre},s_{a},s_{post}]$. For all augmentation methods, we also filter the augmented samples by the basic model (non-aug) to remove the potential noisy samples, as suggested by \cite{CRQDA}.

\subsection{Computation Infrastructure \& Augmentation Efficiency}\label{sec:appendix-speed}
We use 8 NVIDIA V100 cards for \ours{} pre-training and 1 V100 card for all downstream tasks (classification, NER, MRC). In data augmentation process, \ours{} can generate about 700 samples per minute for \texttt{max\_len=60} and about 440 samples per minute for \texttt{max\_len=200} with \texttt{batch\_size=32} on a single V100 card. The speed can be improved by increasing the \texttt{batch\_size} or using more GPUs. Table \ref{augmentation-speed} shows the augmentation speed in different settings.

\begin{table}[h]
\centering
\small
\begin{tabular}{cc|c}
\toprule[1pt]
\multicolumn{3}{c}{\small{\textbf{\ours{} (large)}'s Augmentation Efficiency (single V100)}}   \\
\midrule
\multicolumn{1}{c}{\textbf{{batch size}}} & \multicolumn{1}{c}{\textbf{{max\_len}}} & \multicolumn{1}{|c}{\textbf{{speed (\#/minute)}}} \\
\midrule
32                                      & 60                                              & 698.4                                          \\
32                                      & 200                                             & 444.4                                          \\\midrule
64                                      & 60                                              & 1000.0                                         \\
64                                      & 200                                             & 685.7                              \\
\bottomrule[1pt]
\end{tabular}
\caption{\textbf{\ours{}}'s augmentation efficiency on a single V100, with different settings.}
\label{augmentation-speed}
\end{table}

\section{More Examples}\label{sec:appendix-examples}

\subsection{Sketch-based Generation Examples}\label{sec:more-nlg-examples}
\begin{figure*}
    \centering
    \includegraphics[width=\textwidth]{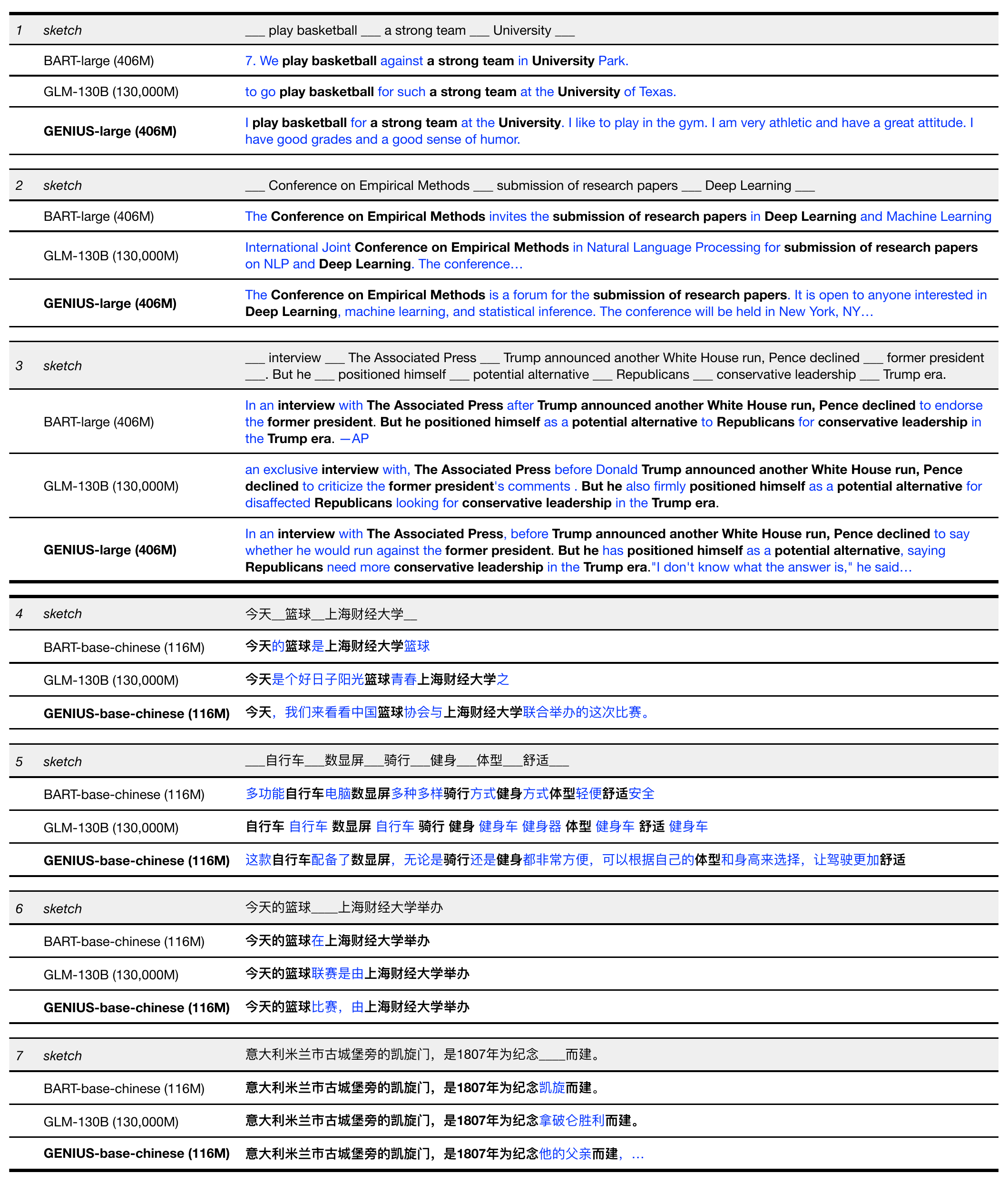}
    \caption{\small{Examples of generated text from BART, GLM-130B and \ours{} in both English and Chinese versions. GLM-130B is a bilingual model with about 130B (130,000M) parameters, while \ours{}-large is English only with about 406M parameters and \ours{}-base-chinese is about 116M parameters, much smaller than the GLM-130B model. We use "\texttt{\_}" to represent the mask token for clearer illustration. The \textbf{bold} are elements from the sketch and the \textcolor{blue}{blue} are newly generated contexts. Sketch 1-8 are typical sketches defined in this work which contain some key information with many masks, while sketch 6-7 are traditional blank-filling tasks where only one mask is waiting for prediction.}}
    \label{fig:compare-glm-130b}
\end{figure*}

We provide some examples to subjectively compare the quality of BART, \ours{} and the giant GLM-130B \cite{zeng2022glm-130b} model in Figure \ref{fig:compare-glm-130b}. GLM-130B is a bilingual model with about 130B (130,000M) parameters pre-trained on over 400 billion text tokens, which is much larger (about 300-1000 times larger) than \ours{} models and uses much more general data for pre-training. Therefore, we expect GLM-130B to be a very strong model for blank-filling tasks. Sketch 1-3 are for English tasks, which show that GLM-130B can also generate fluent sentences given the sketches. The results of GLM-130B are generated by calling their online API\footnote{https://huggingface.co/spaces/THUDM/GLM-130B}. BART can also connect the fragments in a natural way, but can only fill in very few words. \ours{} also handles these inputs well, with even more content inserted than GLM-130B. For the Chinese tasks, we find both BART and GLM-130B generate in-fluent or meaningless text for sketch 4 and 5. Chinese sketch-based reconstruction is more difficult since the tokens are processed at the character level while in English it's word/subword level. However, \ours{} manages to generate coherent text based on these extremely masked sketches. Sketch 6 and 7 are traditional blank-filling tasks with only one mask. All three methods can fill in the blank in a fluent way. However, due to the relatively limited training corpus, BART and \ours{} are unable to give the "precise answer" into the blank in sketch 7, which requires extra knowledge about the concepts described in the context. GLM-130B instead can give the correct answer based on its super large-scale pre-training and model weights.

\subsection{Augmentation Examples}\label{sec:more-aug-examples}
Table \ref{tab:more-exampels-1} gives examples of \ours{}-generated sentences and the effect of using "attribute controlling".

Table \ref{tab:more-examples-clf} illustrates some examples generated by different augmentation methods for text classification tasks. EDA and STA are rule-based methods, while the others are LM-based methods. According to the examples, we can see that \ours{} can generate much more diverse new samples than other methods, while also preserving the core semantics of the original samples.

Table \ref{tab:NER-examples} shows some \ours{}-generated examples for CoNLL03 task. \ours{} introduces new contexts for existing entities. Table \ref{tab:MRC-examples} shows the examples for SQuAD. Based on the target-aware sketch extraction, \ours{} can generate different new paragraphs according to the different questions.

\begin{table*}[t]
\scriptsize
\begin{tabularx}{\linewidth}{cX}
\toprule[1.5pt]
\textit{\textbf{sketch}}        & \texttt{<mask>} \textbf{Conference on Empirical Methods} \texttt{<mask>} \textbf{submission of research papers} \texttt{<mask>} \textbf{Deep Learning} \texttt{<mask>}            \\
\midrule[0.5pt]
\multirow{2}{*}{\textit{\ours{} generated}}& The \textbf{Conference on Empirical Methods} is a forum for the \textbf{submission of research papers} in the field of \textbf{Deep Learning}. The conference is open to all interested parties. It will be held at Stanford University...                  \\
\midrule[1.5pt]
\textit{\textbf{sketch}}        & \texttt{<mask>} \textbf{the European Union} \texttt{<mask>} \textbf{month by EU} \texttt{<mask>} \textbf{Farm Commissioner Franz} \texttt{<mask>}            \\
\midrule[0.5pt]
\multirow{2}{*}{\textit{\ours{} generated}}& Farmers in \textbf{the European Union} will have to pay more for their produce if they want to be included in the EU's new farm bill, which is due to be approved next \textbf{month by EU} lawmakers. EU \textbf{Farm Commissioner Franz} Hoppe said on Wednesday...                  \\
\midrule[0.5pt]
\multirow{2}{*}{\textit{\begin{tabular}[c]{@{}c@{}}prompt=\\ "\textbf{\textcolor{red}{Environment: }}"\end{tabular}}}   & \st{Environment: }What does \textbf{the European Union} need to do? A new report released this \textbf{month by EU}'s \textbf{Farm Commissioner Franz} Hauser says that the EU \textcolor{red}{\textbf{needs to do more to protect the environment}}. He says that there is a ... \\
\midrule[0.5pt]
\multirow{2}{*}{\textit{\begin{tabular}[c]{@{}c@{}}prompt=\\ "\textbf{\textcolor{red}{Technology: }}"\end{tabular}}}    & \st{Technology: }What does it mean for \textbf{the European Union}? A new report released this \textbf{month by EU}'s \textbf{Farm Commissioner Franz} X says that the EU's agricultural sector is in danger of \textcolor{red}{\textbf{losing its competitiveness due to a lack of technological innovation}}. The report... \\
\midrule[0.5pt]
\multirow{2}{*}{\textit{\begin{tabular}[c]{@{}c@{}}prompt=\\ "\textbf{\textcolor{red}{Good News: }}"\end{tabular}}}     & \st{Good News: }The U.S. and \textbf{the European Union} \textcolor{red}{\textbf{are getting closer to a deal}} on farm subsidies. The agreement, which was reached last \textbf{month by EU} Agriculture and \textbf{Farm Commissioner Franz} Haase, will \textcolor{red}{\textbf{give farmers a 1.5 percent increase in their subsidies}}...\\
\midrule[0.5pt]
\multirow{2}{*}{\textit{\begin{tabular}[c]{@{}c@{}}prompt=\\ "\textbf{\textcolor{red}{Bad News: }}"\end{tabular}}}      & \st{Bad News: }\textbf{the European Union} \textcolor{red}{\textbf{will not be able to raise the minimum wage}} to 15 an hour next \textbf{month by EU} standards. \textbf{Farm Commissioner Franz} Haass said that the EU \textcolor{red}{\textbf{would not raise the wage because it was not in the EU's interest}}. The minimum...              \\
\bottomrule[1.5pt]
\end{tabularx}
\caption{Examples of sketches and \ours{} generated samples. The 5-8 rows illustrate the impact of "attribute controlling". After the generation, we delete the prompt text at the beginning. The \textbf{black bold} represents the key parts. The \textcolor{red}{\textbf{red bold}} represents the text that are close to the meaning of the given prompts.}
  \label{tab:more-exampels-1}
\end{table*}

\begin{table*}[t]
\small
\begin{tabularx}{\linewidth}{cX}
\toprule[1.5pt]
\multirow{2}{*}{Original Text}         & I only have eight myself, but I know a few people who have at least 15. They have played a lot of leagues, even in a short amount of years, but in various sports. \textbf{[label: Sports]}           \\
\midrule[1.5pt]
\multirow{2}{*}{\textbf{EDA}}      &    I only have eight myself, but I know a few people who have at least 15. They have played a lot of leagues, even in a short amount of, but in various sports.           \\
\midrule[0.5pt]
\multirow{2}{*}{\textbf{STA}}      &   I only have eight myself but I know few people who have at least 15 They have played a lot of leagues, even in a short amount of years, but in various sports.            \\
\midrule[0.5pt]
\multirow{2}{*}{\textbf{BackTrans}}      &  I only have eight people, but I know a few people who have at least 15. They have played a lot of leagues, even in a few years, but in various sports.             \\
\midrule[0.5pt]
\multirow{2}{*}{\textbf{MLM}}      &    I only have eight myself, but I know a few people who have at least 15. They all played a bunch of leagues, even in a short amount of years, but in various sports           \\
\midrule[0.5pt]
\multirow{2}{*}{\textbf{C-MLM}}      &   \st{\textit{Sports:}} Me only got eight myself, but I have a few buddies who have at least twelve. They have played a lot of leagues, but in a set amount of years time but in various sports.            \\
\midrule[0.5pt]
\multirow{5}{*}{\textbf{LAMBADA}}      &   \st{\textit{Sports:}} I don't think it would be a bad idea to have him on the team.  If he gets hurt, it would certainly hurt him a little bit.  But if he does get hurt, it would hurt him a lot more than any other guy on the team.  If he gets hurt in the playoffs, it would hurt him a lot more than any other guy on the team.  But if he does win, it would hurt him a lot more than any other guy on the team.  And if he does lose, it would hurt him...            \\
\midrule[1.5pt]
\multirow{1}{*}{\textit{sketch}}      &  \textit{Sports:} <mask> lot of leagues <mask> in various sports <mask>             \\
\midrule[0.5pt]
\multirow{1}{*}{\textbf{\oursaug{}}}      &     \st{\textit{Sports:}} There are a lot of leagues in various sports. Here are some of the best.          \\
\midrule[0.5pt]
\multirow{3}{*}{\textbf{\oursaug{}-f}}      &   \st{\textit{Sports:}} I have played in a lot of leagues, in various sports. I'm not into any of them. But I'm a big fan of football. It's a sport that gives you something to do besides playing football. If you want to try something different, try something new.            \\
\midrule[0.5pt]
\multirow{4}{*}{\textit{sketch mixup}}      &    \textit{Sports:} <mask> leagues <mask> High sped <mask> knee and <mask> liverpool rox <mask> leg <mask> various sports <mask> knee <mask> of leagues <mask> knee and straight <mask> Super Bowl XL <mask> sports <mask> rear knee and <mask> lot of leagues <mask> legs <mask> in various sports <mask>           \\
\midrule[0.5pt]
\multirow{5}{*}{\textbf{\ours{}-mixup}}      &    \st{\textit{Sports:}} Which leagues have the High sped up right knee and the liverpool rox right leg? In various sports, the right knee is the most important. In a lot of leagues, the left knee and straight leg are the best. In Super Bowl XL and other sports, it is the right rear knee and right leg that are the most dangerous. In most of the games, the high speed of the leg is the best thing for the player. A lot of teams have the high sped up legs in various sports.           \\
\bottomrule[1.5pt]
\end{tabularx}
\caption{Data augmentation for text classification. During the augmentation, C-MLM, LAMBADA and \oursaug{} all start with a label prompt. The prompt text is removed after generation.}
  \label{tab:more-examples-clf}
\end{table*}

\lstset{
  basicstyle=\ttfamily\scriptsize
  }
  
\begin{table*}[h]
\small
\begin{tabular}{l}
\toprule[1.5pt]
\textbf{Original samples from CoNLL03:}\\
\midrule
\begin{lstlisting}
EU    rejects German call to boycott British lamb .
B-ORG O       B-MISC O    O  O       B-MISC  O    O
\end{lstlisting}\\
\begin{lstlisting}
Germany 's representative to the European Union 's veterinary committee Werner Zwingmann said on Wednesday
B-LOC   O  O              O  O   B-ORG    I-ORG O  O          O         B-PER  I-PER     O    O  O \end{lstlisting}\\
\midrule[1pt]
\textbf{\oursaug{} generated samples:}\\
\midrule
\begin{lstlisting}
The German government says the idea is unacceptable and that the EU    should reject it . 
O   B-MISC O          O    O   O    O  O            O   O    O   B-ORG O      O      O  O 
\end{lstlisting}\\
\midrule
\begin{lstlisting}
The spokesman added that the EU    was not considering imposing a ban on the meat . 
O   O         O     O    O   B-ORG O   O   O           O        O O   O  O   O    O 
\end{lstlisting}\\
\midrule
\begin{lstlisting}
Britain and France are also considering a ban on the sale of tobacco to minors . 
B-LOC   O   B-LOC  O   O    O           O O   O  O   O    O  O       O  O      O  
\end{lstlisting}\\
\midrule
\begin{lstlisting}
The measures have been proposed by EU    Farm and EU    Commissioner Franz Fischler . 
O   O        O    O    O        O  B-ORG O    O   B-ORG O            B-PER I-PER    O 
\end{lstlisting}\\
\midrule
\begin{lstlisting}
`` But Fischler agreed that the EU    s standing on the Spanish Farm is not a problem . 
O  O   B-PER    O      O    O   B-ORG O O        O  O   B-MISC  O    O  O   O O       O 
\end{lstlisting}\\
\midrule
\begin{lstlisting}
It says the meat is being exported to China , Brazil , India and the US .  # unlabeled entity problem
O  O    O   O    O  O     O        O  B-LOC O O      O O     O   O   O  O 
\end{lstlisting}\\
\bottomrule
\end{tabular}
\caption{\oursaug{} generated samples for NER task. The last row shows an example where \ours{} generates new entities like "India", "Brazil" that don't exist in the training data, which results in the unlabeled entity problem. We take the approach in \cite{iclr-ner-negative} to deal with this issue.}
\label{tab:NER-examples}
\end{table*}

\begin{table*}[h]
\small
\begin{tabularx}{\linewidth}{X}
\toprule[1.5pt]
\multicolumn{1}{l}{\textbf{Original sample from SQuAD:}}\\
\textbf{Paragraph:} Architecturally, the school has a Catholic character. Atop the Main Building's gold dome is a golden statue of the Virgin Mary. Immediately in front of the Main Building and facing it, is a copper statue of Christ with arms upraised with the legend "Venite Ad Me Omnes". Next to the Main Building is the Basilica of the Sacred Heart. Immediately behind the basilica is the Grotto, a Marian place of prayer and reflection. \textit{It is a replica of the grotto at Lourdes, France where the Virgin Mary reputedly appeared to \underline{Saint Bernadette Soubirous} in 1858.} At the end of the main drive (and in a direct line that connects through 3 statues and the Gold Dome), is a simple, modern stone statue of Mary. \\
\textbf{Question:} To whom did the Virgin Mary allegedly appear in 1858 in Lourdes France?\\
\midrule[1pt]
\multicolumn{1}{l}{\textbf{New paragraph generated by \oursaug{}:}}\\
\textbf{Paragraph:} The basilica has a golden statue of the Virgin Mary and a statue of Christ, which is hidden away behind the basilica. \textit{It is a replica of the grotto at Lourdes, France where the virgin Mary reputedly appeared to \underline{Saint Bernadette Soubirous} in 1858.} The basilica contains 3 statues and a modern stone statue of Mary, which was built in the late 1800s. There are also a number of other relics. The relics are kept in a special room on the main floor of the church. The church is open to the public from 9:00am to 4:00pm Monday  Friday and 9:30am to 3:00 pm on Saturday and Sunday. The museum is open from 10:00 am to 5:30 pm on weekdays. \\
\midrule[1.5pt]
\multicolumn{1}{l}{\textbf{Original sample from SQuAD (same paragraph with a \textit{different} question):}}\\
\textbf{Paragraph:} Architecturally, the school has a Catholic character. Atop the Main Building's gold dome is a golden statue of the Virgin Mary. \textit{Immediately in front of the Main Building and facing it, is \underline{a copper statue of Christ} with arms upraised with the legend "Venite Ad Me Omnes".} Next to the Main Building is the Basilica of the Sacred Heart. Immediately behind the basilica is the Grotto, a Marian place of prayer and reflection. It is a replica of the grotto at Lourdes, France where the Virgin Mary reputedly appeared to Saint Bernadette Soubirous in 1858. At the end of the main drive (and in a direct line that connects through 3 statues and the Gold Dome), is a simple, modern stone statue of Mary. \\
\textbf{Question:} What is in front of the Notre Dame Main Building?\\
\midrule[1pt]
\multicolumn{1}{l}{\textbf{New paragraph generated by \oursaug{}:}}\\
\textbf{Paragraph:} In front of the school and the Main Building is a gold dome and a golden statue of the Madonna. \textit{Immediately in front of our Main Building and facing it, is \underline{a copper statue of Christ} with arms upraised with the legend "Venite Ad Me Omnes" on the top of the dome.} In the Basilica of the Immaculate Conception, there is a small chapel with a small altar. Also behind the basilica is a grotto at Lourdes. There is a statue of Mary in the center of the grotto. We recommend that you take a bus to the nearest bus stop, as there is ample parking. We suggest that you arrive early in the morning to avoid long lines.\\
\bottomrule
\end{tabularx}
\caption{\oursaug{} generated samples for SQuAD. Given a paragraph, \oursaug{} can generate different new paragraphs according to the given question, based on the target-aware sketch extraction. The \underline{underlined} words are the ground truth answers, the \textit{italicized} sentences are the key sentences that contain the answer, which are kept in the sketch during augmentation. }
\label{tab:MRC-examples}
\end{table*}

\end{document}